\documentclass{article}

     \usepackage[final]{neurips_2020}

\usepackage[utf8]{inputenc} % allow utf-8 input
\usepackage[T1]{fontenc}    % use 8-bit T1 fonts
\usepackage{hyperref}       % hyperlinks
\usepackage{url}            % simple URL typesetting
\usepackage{booktabs}       % professional-quality tables
\usepackage{amsfonts}       % blackboard math symbols
\usepackage{nicefrac}       % compact symbols for 1/2, etc.
\usepackage{microtype}      % microtypography
\usepackage{xcolor}
\usepackage{amsmath}
\usepackage{graphicx}
\usepackage{wrapfig}
\usepackage[ruled,vlined]{algorithm2e}
\SetKwComment{Comment}{$\triangleright$\ }{}
\SetCommentSty{itshape}
\usepackage{amsthm}
\usepackage{subfig}
\usepackage{array,multirow}
\usepackage{wrapfig}
\usepackage{changepage}

\theoremstyle{definition}
\newtheorem{defn}{Definition}
\newtheorem*{defn*}{Definition}
\newtheorem*{theorem*}{Theorem}
\newtheorem{theorem}{Theorem}

\theoremstyle{remark}
\newtheorem*{argument}{Argument}

\newboolean{draftmode}
\setboolean{draftmode}{false}

\usepackage{color}
\newcommand{\mycomment}[3]{{\textcolor{#3}{#1 #2}}}
\newcommand{\pfaumarker}[1]{\textcolor{blue}{\emph{Pfau:} #1}}
\newcommand{\todopfaumarker}[1]{\textcolor{blue}{\emph{TODO(Pfau):} #1}}
\newcommand{\todosebmarker}[1]{\textcolor{red}{\emph{TODO(Seb):} #1}}
\newcommand{\sebmarker}[1]{\textcolor{red}{\emph{Seb:} #1}}
\newcommand{\ihmarker}[1]{\textcolor{darkgreen}{\emph{Irina:} #1}}
\newcommand{\todoihmarker}[1]{\textcolor{darkgreen}{\emph{TODO(Irina):} #1}}
\newcommand{\abmarker}[1]{\textcolor{purple}{\emph{Alex:} #1}}
\newcommand{\drewmarker}[1]{\textcolor{orange}{\emph{Drew:} #1}}

\ifthenelse{\boolean{draftmode}}{
 \newcommand{\pfau}[1]{\mycomment{\pfaumarker}{#1}{blue}}
 \newcommand{\todopfau}[1]{\mycomment{\todopfaumarker}{#1}{blue}}
 \newcommand{\todoseb}[1]{\mycomment{\todosebmarker}{#1}{red}} 
 \newcommand{\seb}[1]{\mycomment{\sebmarker}{#1}{red}} 
 \newcommand{\ih}[1]{\mycomment{\ihmarker}{#1}{darkgreen}} 
 \newcommand{\todoih}[1]{\mycomment{\todoihmarker}{#1}{darkgreen}}
  \newcommand{\ab}[1]{\mycomment{\abmarker}{#1}{purple}}
  \newcommand{\drew}[1]{\mycomment{\drewmarker}{#1}{orange}}
}{
 \newcommand{\pfau}[1]{}
 \newcommand{\todopfau}[1]{}
 \newcommand{\todoseb}[1]{}
 \newcommand{\seb}[1]{}
 \newcommand{\ih}[1]{}
 \newcommand{\todoih}[1]{}
 \newcommand{\ab}[1]{} 
 \newcommand{\drew}[1]{} 
}

\usepackage{xspace}
\usepackage[subtle,tracking=normal]{savetrees}

\newcommand{\geomancer}[0]{\textsc{GeoManCEr}}
\newcommand{\laplsym}[0]{\overline{\Delta^2}}

\definecolor{darkgreen}{cmyk}{1.0, 0, 1.0, 0}
\definecolor{orange}{cmyk}{0, 0.5, 1, 0}
\definecolor{purple}{cmyk}{0.5, 1, 0, 0}

\title{Disentangling by Subspace Diffusion}

\author{%
  David~Pfau, Irina Higgins, Aleksandar Botev, S{\'e}bastien Racani\`ere \\
  DeepMind\\
  London, UK \\
  \texttt{\{pfau, irinah, botev, sracaniere\}@google.com} \\
}

\begin{document}

\maketitle

\begin{abstract}
  We present a novel nonparametric algorithm for symmetry-based disentangling of data manifolds, the Geometric Manifold Component Estimator (\geomancer). \geomancer\, provides a partial answer to the question posed by Higgins et al. (2018): is it possible to learn how to factorize a Lie group solely from observations of the orbit of an object it acts on? We show that fully unsupervised factorization of a data manifold is possible {\em if} the true metric of the manifold is known and each factor manifold has nontrivial holonomy -- for example, rotation in 3D. Our algorithm works by estimating the subspaces that are invariant under random walk diffusion, giving an approximation to the de Rham decomposition from differential geometry. We demonstrate the efficacy of \geomancer\, on several complex synthetic manifolds. Our work reduces the question of whether unsupervised disentangling is possible to the question of whether unsupervised metric learning is possible, providing a unifying insight into the geometric nature of representation learning.
\end{abstract}

\section{Introduction}

The ability to disentangle the different latent factors of variation in the world has been hypothesized as a critical ingredient in representation learning \citep{bengio2013representation}, and much recent research has sought a fully unsupervised way to learn disentangled representations from data \cite{ridgeway2018learning,chen2016infogan,chen2018isolating,dupont2018learning,kim2018disentangling,esmaeili2018structured,kumar2018variational,achille2018lifelong,ansari2019hyperprior,matthieu2019disentangling,detlefsen2019explicit,dezfouli2019behavioural,lorenz2019partbased,lee2020idgan,casellesdupre2019disentangled,quessard2020learning,ramesh2019spectral}. Because the community has not settled on a definition of ``disentangling", much of this work relies on heuristics and qualitative criteria to judge performance. For instance, datasets are often constructed by varying interpretable factors like object position, rotation and color % position in $x$ and $y$, object color, and rotation, 
and methods are judged by how well they recover these predefined factors \cite{higgins2017beta,locatello2019challenging}. One commonly used definition is that a representation is disentangled if the data distribution can be modeled as a nonlinear transformation of a product of independent probability distributions \citep{locatello2019challenging}. This leads to a pessimistic result, that the different latent factors are not identifiable without side information or further assumptions.

To escape this pessimistic result, we can turn to a different, symmetry-based definition of disentangling \citep{higgins2018towards}, rooted in the Lie group model of visual perception \citep{hoffman1966lie, dodwell1983lie, rao1999learning, sohl2010unsupervised, anselmi2016invariance}. Instead of a product of {\em distributions}, the symmetry-based approach to disentangling considers a representation disentangled if it matches the product of {\em groups} that define the symmetries in the world. If the actions that define the possible transformations in the world form a group $G=G_1 \times G_2 \times \ldots \times G_m$, then a representation is disentangled under this definition if it decomposes in such a way that the action of a single subgroup leaves all factors of the representation invariant except for one (See Supp. Mat., Sec.~\ref{sec:proofs} for a formal definition).

%Rather than drawing on probabilistic machine learning, this approach is rooted in the Lie group model of visual perception \citep{hoffman1966lie, dodwell1983lie, rao1999learning, sohl2010unsupervised, anselmi2016invariance}, which focuses on the continuous transformations that leave objects invariant, such as changes to pose and illumination. Group theory is a natural way to describe symmetries that exist in the world: every element of a group is one transformation that can be applied to an object, and transformations can be composed or inverted. Instead of a product of {\em distributions}, the symmetry-based approach to disentangling considers a representation disentangled if it matches the product of {\em groups} that define the symmetries in the world. If the actions that define the possible transformations in the world form a group $G=G_1 \times G_2 \times \ldots \times G_m$, then a representation is disentangled under this definition if it decomposes in such a way that the action of a single subgroup leaves all factors of the representation invariant except for one (See Supp. Mat., Sec.~\ref{sec:proofs} for a formal definition).

\begin{figure}
    \centering
    \includegraphics[width=0.9\textwidth]{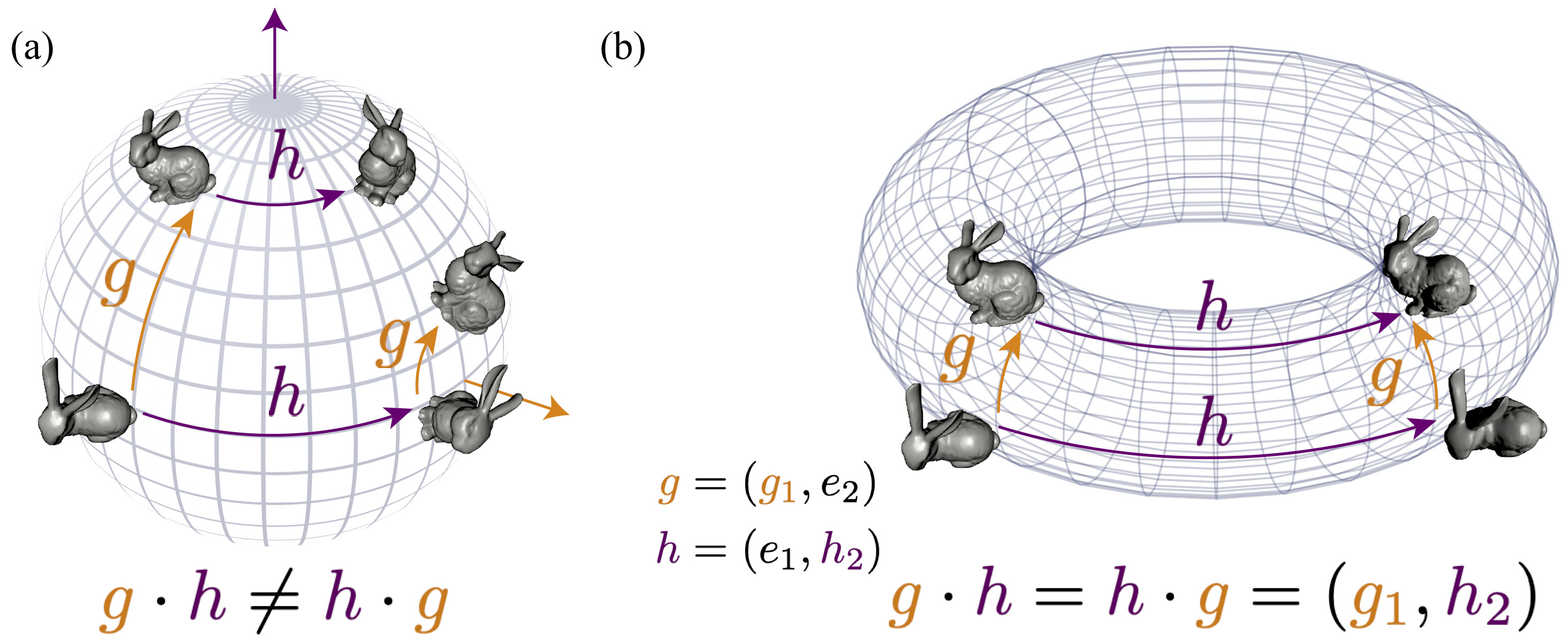}
    \caption{Two views of the orbit of an object under the action of a group, in this case images of the Stanford Bunny under changes in pose and illumination (a) Transformations from the same subgroup, in this case 3D rotations, do not in general commute, and the analogy is ambiguous. (b) Transformations from disentangled subgroups. The transformations commute and the analogy is unambiguous.}
    \label{fig:analogy}
    \vspace{-10pt}
\end{figure}

The symmetry-based definition is appealing as it resolves an apparent contradiction in the parallelogram model of analogical reasoning \citep{rumelhart1973model}. For concepts represented as vectors $\mathbf{a}$, $\mathbf{b}$ and $\mathbf{c}$ in a flat space, the analogy $\mathbf{a}:\mathbf{b}::\mathbf{c}:\mathbf{d}$ can be completed by $\mathbf{d} = \mathbf{b} + \mathbf{c} - \mathbf{a}$, as $\mathbf{d}$ completes the fourth corner of a parallelogram. This model has worked well in practice when applied to embeddings learned by deep neural networks for words \citep{mikolov2013efficient, pennington2014glove} and natural images \citep{reed2015deep, radford2016unsupervised}, and often matches human judgments \citep {chen2017evaluating}. But for many natural transformations, including 3D rotation, it is not possible to form a representation in a flat vector space such that vector addition corresponds to composition of transformations. Instead, observations from such transformations can be naturally represented as coming from the orbit of a group \cite{anselmi2016invariance}. In this setting, the parallelogram model breaks down -- if $a, b, c$ are representations, and $g, h\in G$ are the group elements such that $g \cdot a = b$ and $h \cdot a = c$, then the completion of the analogy $a:b::c:d$ is ambiguous, as $g \cdot h \cdot a \ne h \cdot g \cdot a$ for noncommutative operations (Figure~\ref{fig:analogy}). State-of-the-art generative models for single image classes typically elide this by restricting the dataset to a limited patch of the manifold, like forward-facing faces \citep{karras2019analyzing} or limited ranges of change in elevation \cite{lecun2004learning, liu2015faceattributes, fidler20123d, kim2018disentangling, aubry2014seeing, paysan20093d}. This ambiguity, however, is resolved if $G$ is a product of subgroups. So long as $g$ and $h$ leave all factors invariant except for one, and each varies a different factor, then $g$ and $h$ do commute, and the analogy can be uniquely completed. 

%This problem can be avoided by representing observations as coming from the orbit of a group \cite{anselmi2016invariance}. In this setting, the parallelogram model breaks down -- if $a, b, c$ are representations, and $g, h\in G$ are the group elements such that $g \cdot a = b$ and $h \cdot a = c$, then the completion of the analogy $a:b::c:d$ is ambiguous, as $g \cdot h \cdot a \ne h \cdot g \cdot a$ for noncommutative operations (Figure~\ref{fig:analogy}). Even state-of-the-art generative models for single image classes typically elide this by restricting the dataset to a limited patch of the manifold, like forward-facing faces \citep{karras2019analyzing} or limited ranges of change in elevation \cite{lecun2004learning, liu2015faceattributes, fidler20123d, kim2018disentangling, aubry2014seeing, paysan20093d}. This ambiguity can be resolved if $G$ is a product of subgroups. So long as $g$ and $h$ leave all factors invariant except for one, and each varies a different factor, then $g$ and $h$ do commute, and the analogy can be uniquely completed.

While this definition of disentangling is appealing, it does not provide an obvious recipe for how to learn the appropriate factorization of the action on the world state. Some group invariances and equivariances can be built into neural network architectures \citep{fukushima1980neocognitron, gens2014deep, cohen2016group, cohen2018spherical, kondor2018clebsch, weiler20183d, keriven2019universal, finzi2020generalizing}, and it has been shown how commutative group representations can be learned \citep{cohen2014learning}. Methods have been proposed to learn symmetry-based disentangled representations when conditioned on interactions \cite{casellesdupre2019disentangled, quessard2020learning}, or to learn dictionaries of Lie group operators from neighboring pairs of data \cite{culpepper2009learning, connor2020representing}, but a general algorithm to factorize noncommutative groups without any supervision remains elusive.

If we restrict our attention to Lie groups -- groups that are also manifolds, like rotations -- we could use the properties of infinitesimal transformations as a learning signal. Essentially, we would like to use failures of the parallelogram model {\em as a learning signal itself}. %Those directions that are disentangled from one another will be those that behave like vectors in a flat space when one is moved in the direction of the other, while directions from the same entangled submanifold may be mixed together in arbitrary ways \drew{Explicitly connect this sentence to the point about using the parallelogram rule? I find it slightly ungainly as is}. 
Those directions that lie on disentangled submanifolds will behave like vectors in a flat space when one is moved in the direction of the other, hence complying with the parallelogram model, while directions within each submanifold may be mixed together in arbitrary ways. Computing all of the directions that remain invariant provides a disentangled factorization of the manifold. These intuitions can be made precise by the de Rham decomposition theorem \cite{derham1952reductibilite}, a foundational theorem in holonomy theory, a branch of differential geometry.

Here we present an algorithm that turns these ideas into a practical method for disentangling, the Geometric Manifold Component Estimator (\geomancer). \geomancer\, differs from other disentangling algorithms in that it does not learn a nonlinear embedding of the data. Instead, it can either be applied directly to the data so long as the local metric information is known, or it can be applied as a post-processing step to learned embeddings. \geomancer\, is a nonparametric algorithm which learns a set of subspaces to assign to each point in the dataset, where each subspace is the tangent space of one disentangled submanifold. This means that \geomancer\, can be used to disentangle manifolds for which there may not be a global axis-aligned coordinate system. \geomancer\, is also able to discover the correct number of dimensions in each submanifold without prior knowledge. Our algorithm is particularly well suited for dealing with transformations which have nontrivial holonomy, such as 3D rotations. In contrast, most previous work \cite{ridgeway2018learning,chen2016infogan,chen2018isolating,dupont2018learning,kim2018disentangling,esmaeili2018structured,kumar2018variational,achille2018lifelong,ansari2019hyperprior,matthieu2019disentangling,detlefsen2019explicit,dezfouli2019behavioural,lorenz2019partbased,lee2020idgan,casellesdupre2019disentangled,ramesh2019spectral} has focused on transformations with trivial holonomy, such as translation in 2D.  %In contrast, previous work arguing that interactions are necessary to disentangle transformations \cite{casellesdupre2019disentangled} has focused on transformations with trivial holonomy, such as translation in 2D. 

\geomancer\, builds on classic work on nonparametric manifold learning, especially Laplacian Eigenmaps \cite{belkin2002laplacian}, Diffusion Maps \cite{coifman2006diffusion} and extensions like Orientable and Vector Diffusion Maps \cite{singer2011orientability, singer2012vector, fan2019multi}, generalizing the idea of finding modes of a random walk diffusion operator on manifolds from scalars and vectors to matrices. These classic methods are primarily focused on learning a low-dimensional embedding of data. While \geomancer\, uses and extends many of these same mathematical techniques, its primary goal is to learn a factorization of the data manifold, which makes it an interesting new application of spectral graph theory.

 \geomancer\, should not be confused with methods like Horizontal Diffusion Maps, which are for ``synchronization" problems in manifold learning \cite{gao2019geometry, gao2019diffusion}. These methods also use scalar and (co)vector diffusion, but not subspace diffusion. Much like other manifold learning problems, they are concerned with finding an embedding of points on a manifold, not a factorization. What distinguishes synchronization problems from the rest of manifold learning is that they are concerned with finding embeddings such that operations that relate pairs of data can be composed together in a cycle-consistent manner. \geomancer\, instead exploits the ``cycle-inconsistency" of certain operations to distinguish entangled and disentangled directions around each point.

It is important to distinguish \geomancer\, from other manifold learning methods that go beyond learning a single undifferentiated manifold, especially Robust Multiple Manifold Structure Learning (RMMSL) \cite{gong2012robust}. RMMSL learns a {\em mixture} of manifolds which are all embedded in the same space. \geomancer\, by contrast, learns a single manifold which is itself a {\em product} of many submanifolds, where each submanifold exists in its own space. We next present a rapid overview of the relevant theory, followed by a detailed description of \geomancer, and finally show results on complex data manifolds.

\section{Theory}
We assume some basic familiarity with the fundamentals of Riemannian geometry and parallel transport, though we strongly encourage reading the review in Supp. Mat., Sec.~\ref{sec:diff_geom_review}. For a more thorough treatment, we recommend the textbooks by Do Carmo \cite{do2016differential} and Kobayashi and Nomizu \cite{kobayashi1963foundations}. Let $x$ denote points on the $k$-dimensional Riemannian manifold $\mathcal{M}$ with metric $\langle \cdot, \cdot \rangle_\mathcal{M}$, possibly embedded in $\mathbb{R}^n$. We denote paths by $\gamma:\mathbb{R}\rightarrow\mathcal{M}$, tangent spaces of velocity vectors by $T_x\mathcal{M}$, cotangent spaces of gradient operators by $T_x^*\mathcal{M}$, and vectors in $T_x\mathcal{M}$ by $\mathbf{v}$, $\mathbf{w}$, etc. As $\mathcal{M}$ is a Riemannian manifold, it has a unique Levi-Civita connection, which allows us to define a covariant derivative and parallel transport, which gives a formal definition of two vectors in nearby tangent spaces being parallel.

\begin{wrapfigure}{r}{7cm}
    \vspace{-29pt}
    \centering
    \includegraphics[width=0.35\textwidth]{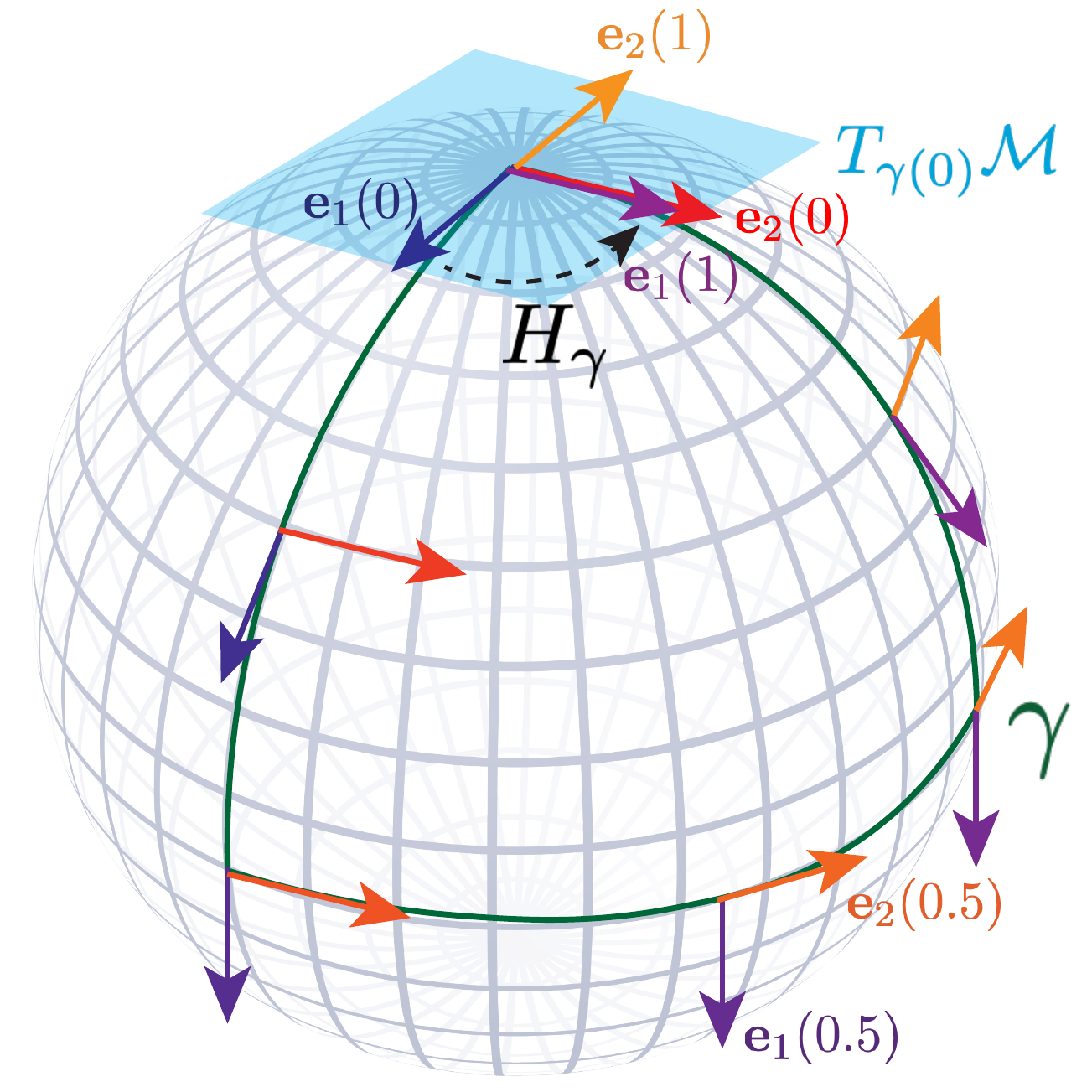}
    \caption{Holonomy on the sphere. The linear transform $H_\gamma$ captures the amount vectors in $T_{\gamma(0)}\mathcal{M}$ are rotated by parallel transport around the path $\gamma$.}
    \label{fig:holonomy}
\end{wrapfigure}

%, where $\mathrm{tr}\left(\laplsym[\mathbf{\Sigma}]\right)=0$. If the diffusion is ergodic, $\mathbf{\Sigma}(t)$ converges to $\mathrm{tr}\left(\mathbf{\Sigma}(0)\right)\mathbf{I}$ everywhere.

\paragraph{Holonomy and the de Rham decomposition}
The basic tools of differential geometry, especially parallel transport, can be used to infer the product structure of the manifold through the {\em holonomy group}. Consider a loop $\gamma:[0, 1]\rightarrow \mathcal{M}$, $\gamma(0)=x$, $\gamma(1)=x$. Given an orthonormal basis $\{\mathbf{e}_i\}\subset T_x\mathcal{M}$ of the tangent space at $x$, we can consider the parallel transport of all basis vectors around the loop $\{\mathbf{e}_i(t)\}$, illustrated in Fig.~\ref{fig:holonomy}. Then the vectors $\{\mathbf{e}_i(1)\}\subset T_x\mathcal{M}$ form the columns of a matrix $H_\gamma$ that fully characterizes how any vector transforms when parallel transported around a loop. That is, for any parallel transport $\mathbf{v}(t)$ along $\gamma$, $\mathbf{v}(1) = H_\gamma \mathbf{v}(0)$. Note that if the affine connection preserves the metric \seb{Don't you assume that the connection is the Levi-Civita connection everywhere else? In which case you can get rid of this if statement and directly say above that $H_\gamma$ is an orthogonal matrix.}, then $\{\mathbf{e}_i(t)\}$ will be an orthonormal basis for all $t$, so $H_\gamma$ is in the orthonormal group $\mathrm{O}(k)$. Moreover if $\mathcal{M}$ is orientable, then the handedness of $\{\mathbf{e}_i(t)\}$ cannot change, so $H_\gamma$ is in the special orthonormal group $\mathrm{SO}(k)$, that is, $\mathrm{det}\left(H_\gamma\right)=1$.\seb{You could shorten that by not-mentionning handedness and just saying: Moreover if $\mathcal{M}$ is orientable then $H_\gamma$ is in $\mathrm{SO}(k)$.}

% \begin{figure}
%     \centering
%     \subfloat[][]{
%     \includegraphics[width=0.35\textwidth]{fig2.pdf}
%     \label{fig:holonomy}
%     }\hspace{2cm}
%     \subfloat[][]{
%     \includegraphics[width=0.35\textwidth]{fig3_small.pdf}
%     \label{fig:diffusion}
%     }
%     \caption{Illustration of parallel transport, holonomy and subspace diffusion. (a) Holonomy on the sphere. The linear transform $H_\gamma$ captures the amount vectors in $T_{\gamma(0)}\mathcal{M}$ are rotated by parallel transport around the path $\gamma$. (b) Subspace diffusion. A field of symmetric semidefinite matrices $\mathbf{\Sigma}(t)$ evolves in time according to the differential equation $\dot{\mathbf{\Sigma}} = \laplsym[\mathbf{\Sigma}]$. Those directions which either collapse inward or expand outward the slowest (corresponding to the projection matrices $\mathbf{\Pi}_x^-$ and $\mathbf{\Pi}_x^+$) are the subspaces closest to being invariant under parallel transport.}
%     \label{fig:holonomy_and_subspace_diffusion}
%     \vspace{-10pt}
% \end{figure}

The linear transform $H_\gamma$ is the {\em holonomy} of the loop $\gamma$. The space of all holonomies for all loops that start and end at $x$ form the {\em holonomy group} $\mathrm{Hol}_x(\mathcal{M})$ at $x$. \ih{You could move the rest of the paragraph into the appendix}\seb{+1. That would save quite a bit of space.} It can clearly be seen that this space is a group by considering different loops. The trivial loop $\gamma(t) = x$ has holonomy $I$, the identity. If $\gamma$ has holonomy $H_\gamma$, then the loop $\gamma'(t) = \gamma(1-t)$ has holonomy $H_\gamma^{-1}$, so if $H_\gamma$ is in the holonomy group, so is its inverse. And if $\gamma_1$ and $\gamma_2$ are loops, then the loop formed by first going around $\gamma_1$ followed by $\gamma_2$ has holonomy $H_{\gamma_2} H_{\gamma_1}$, so if two elements are in this group, then so is their product.

The structure of the holonomy group is extremely informative about the global structure of the manifold. If the manifold $\mathcal{M}$ is actually a product of submanifolds $\mathcal{M}_1 \times \mathcal{M}_2 \times \ldots \times \mathcal{M}_n$, with the corresponding product metric as its metric \seb{You could delete "as its metric"}, then it is straightforward to show that \seb{You could also delete "it is straightforward to show that"} the tangent space $T_x\mathcal{M}$ can be decomposed into orthogonal subspaces $T^{(1)}_x\mathcal{M},\ldots, T^{(m)}_x\mathcal{M}$ such that the action of $\mathrm{Hol}_x(\mathcal{M})$ leaves each subspace invariant. That is, if $\mathbf{v}\in T^{(i)}_x\mathcal{M}$, then $H_\gamma \mathbf{v} \in T^{(i)}_x\mathcal{M}$ for all $\gamma$ \seb{I would also suggest deleting this sentence. You've already said its content using words instead of symbols.}. These subspaces are each tangent to the respective submanifolds that make up $\mathcal{M}$. The more remarkable result is that the converse holds locally and, if the manifold is simply connected and geodesically complete, globally \citep{derham1952reductibilite}. A manifold is simply connected if any closed loop can be continuously deformed into a single point, and it is geodesically complete if any geodesic can be followed indefinitely.

\begin{theorem}{de Rham Decomposition Theorem (de Rham, 1952), see also \cite[Theorem 6.1]{kobayashi1963foundations}:}
Assume $\mathcal{M}$ is a simply connected and geodesically complete Riemannian manifold. If there exists a point $x\in\mathcal{M}$ and a proper subspace $U$ that is invariant under the action of the holonomy group $\mathrm{Hol}_x(\mathcal{M})$, then $\mathcal{M}$ is a product Riemannian manifold $\mathcal{M}_1\times\mathcal{M}_2$ with $T_x\mathcal{M}_1=U$ and $T_x\mathcal{M}_2=U^\perp$. The tangent spaces to $\mathcal{M}_1$ and $\mathcal{M}_2$ at any other point $y$ are obtained by parallel transporting $U$ and $U^\perp$ respectively along any path from $x$ to $y$.
\end{theorem}

The above theorem can be applied recursively, so that if the holonomy group leaves multiple pairwise orthogonal subspaces invariant, we can conclude that $\mathcal{M}$ is a product of multiple Riemannian manifolds. It seems quite remarkable that a property of the holonomy group in a single tangent space can tell us so much about the structure of the manifold. This is because the holonomy group itself integrates information over the entire manifold, so in a sense it is not really a local property at all.

This result is the main motivation for \geomancer\, -- we aim to discover a decomposition of a data manifold by investigating its holonomy group. The holonomy group is a property of {\em all possible paths}, so it cannot be computed directly. Instead, we build a partial differential equation that inherits properties of the holonomy group and work with a numerically tractable approximation to this PDE.

\begin{adjustwidth}{5.3cm}{0cm}
\nointerlineskip\leavevmode
\paragraph{Subspace diffusion on manifolds}

While it is not feasible to compute properties of all loops on a manifold, the {\em average} properties of random walk diffusion on a manifold can be computed by studying the properties of the diffusion equation. Consider a particle undergoing a Brownian random walk on a manifold with diffusion rate $\tau$. Then, given an initial probability density $p(x, 0)$, the probability of finding the particle at $x$ at time $t$ evolves according to the diffusion equation:
\end{adjustwidth}

\begin{equation}
    \hspace{6cm}\frac{\partial p(x, t)}{\partial t} = \tau \Delta^0[p](x, t)
    \label{eqn:diffusion}
\end{equation}

\begin{wrapfigure}{l}{5cm}
    \vspace{-160pt}
    \centering
    \includegraphics[width=0.35\textwidth]{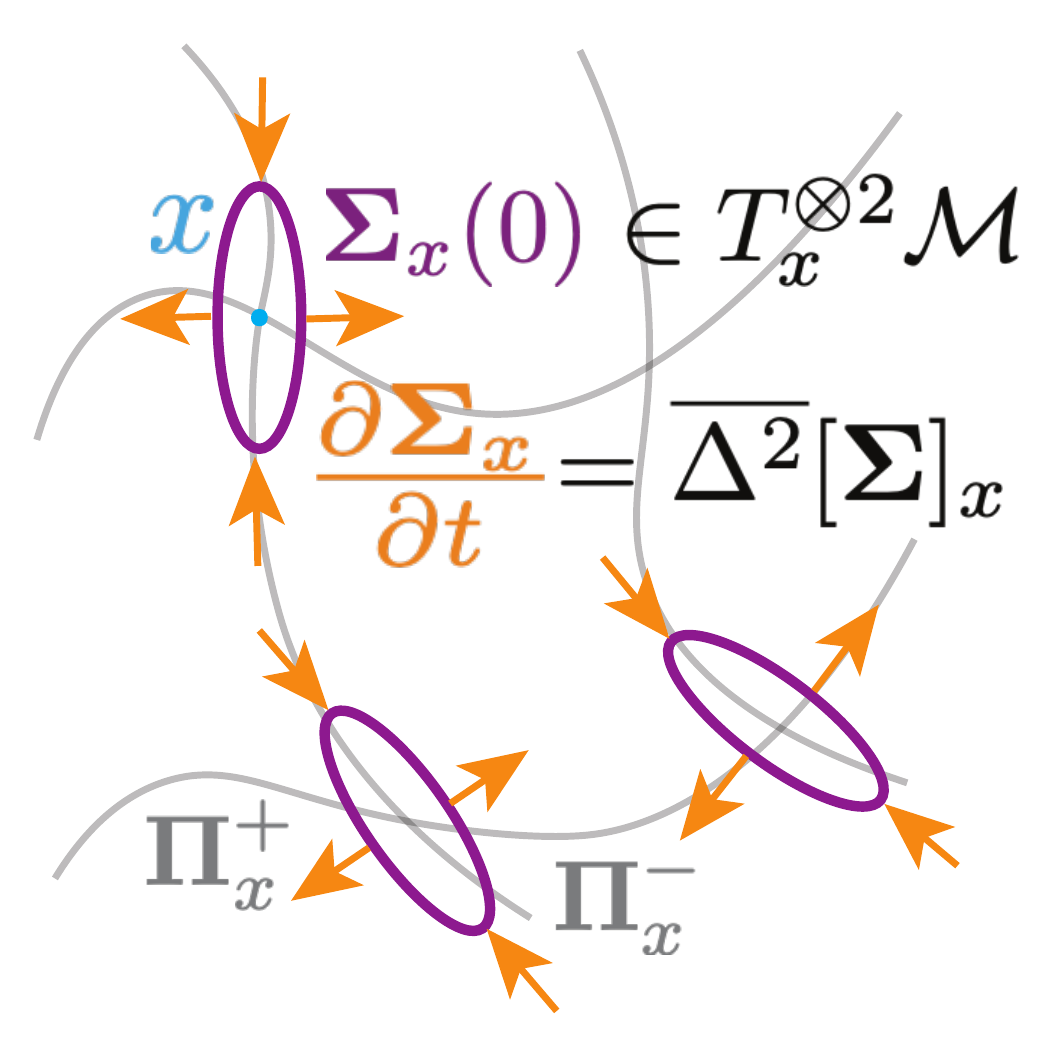}
    \caption{Subspace diffusion. A field of symmetric semidefinite matrices $\mathbf{\Sigma}(t)$ evolves in time according to the differential equation $\dot{\mathbf{\Sigma}} = \laplsym[\mathbf{\Sigma}]$.}
    \label{fig:diffusion}
    \vspace{-10pt}
\end{wrapfigure}

where $\Delta^0$ is a linear operator called the {\em Laplace-Beltrami} operator \cite[Note 14, Vol. 2]{kobayashi1963foundations}, defined as the trace of the second covariant derivative $\Delta^0[f] = \mathrm{Tr}\nabla^2 f$. Even if the initial condition is a delta function, the change in probability is nonzero everywhere, so the Laplace-Beltrami operator encodes global information about the manifold, though it weights local information more heavily.

The Laplace-Beltrami operator, which acts on scalar functions, can be generalized to the {\em connection Laplacian} for rank-$(p, q)$ tensor-valued functions\seb{We could simply call that "rank-$(p, q)$ tensors" instead of "tensor-value functions"}. As the second covariant derivative of a rank-$(p,q)$ tensor is a rank-$(p,q+2)$ tensor, we can take the trace over the last two dimensions to get the connection Laplacian. The connection Laplacian also has an intuitive interpretation in terms of random walks of vectors. Given a probability density $p(\mathbf{v}, t)$ over $T\mathcal{M}$\seb{This is getting very technical, but to talk about a density on $T\mathcal{M}$, we need to have a measure on that space. The measure here comes from a metric on $T\mathcal{M}$. So far we have introduced a metric on $M$, not on $T\mathcal{M}$. You can get such a metric from the metric on $M$, but I don't know if this is obvious or trivial.}, the manifold of all tangent spaces on $\mathcal{M}$ \seb{You have already used $T\mathcal{M}$, so you don't need to redefine it.}, the connection Laplacian on vectors $\Delta^1$ gives the rate of change of the mean of the density at every point $\mu_x = \int_{T_x\mathcal{M}} p(\mathbf{v}, t) \mathbf{v} d\mathbf{v}$, while the connection Laplacian on matrices $\Delta^2$ gives the rate of change of the second moment $\mathbf{\Sigma}_x = \int_{T_x\mathcal{M}} p(\mathbf{v}, t)\mathbf{v}\mathbf{v}^T d\mathbf{v}$ (Fig.~\ref{fig:diffusion}).

Many of the properties of the holonomy group can be inferred from the second-order connection Laplacian $\Delta^2$. In particular, for a product manifold, the eigenfunctions of $\Delta^2$ contain information about the invariant subspaces of the holonomy group:

\begin{theorem}{}
Let $\mathcal{M} = \mathcal{M}_1 \times \ldots \times \mathcal{M}_m$ be a Riemannian product manifold, and let $T^{(1)}_x\mathcal{M},\ldots,T^{(m)}_x\mathcal{M}$ denote orthogonal subspaces of $T_x\mathcal{M}$ that are tangent to each submanifold. Then the tensor fields $\mathbf{\Pi}^{(i)}:\mathcal{M}\rightarrow T_x\mathcal{M}\otimes T^*_x\mathcal{M}$ for $i \in 1, \ldots, m$, where $\mathbf{\Pi}^{(i)}_x$ is the linear projection operator from $T_x\mathcal{M}\rightarrow T^{(i)}_x\mathcal{M}$, go to 0 under the action of the connection Laplacian $\Delta^2$.
\end{theorem}
We provide an informal argument and a more formal proof in Supp. Mat., Sec.~\ref{sec:proofs}.
There is an elegant parallel with the scalar Laplacian. The number of zero eigenvalues of the Laplacian is equal to the number of connected components of a graph or manifold, with each eigenvector being uniform on one component and zero everywhere else. For the second-order Laplacian, the zero eigenvalues correspond to factors of a product manifold, with the matrix-valued eigenfunction being the identity in the subspace tangent to one manifold and zero everywhere else. However, these are not in general the {\em only} eigenfunctions of $\Delta^2$ with zero eigenvalue, and we will discuss in the next section how \geomancer\, avoids spurious eigenfunctions.

\section{Method}
Here we describe the actual Geometric Manifold Component Estimator algorithm (\geomancer). The main idea is to approximate the second-order connection Laplacian $\Delta^2$ from finite samples of points on the manifold, and then find those eigenvectors with nearly zero eigenvalue that correspond to the disentangled submanifolds of the data. This allows us to define a set of {\em local} coordinates around every data point that are aligned with the disentangled manifolds.

\begin{figure}
    \centering
    \includegraphics[width=\textwidth]{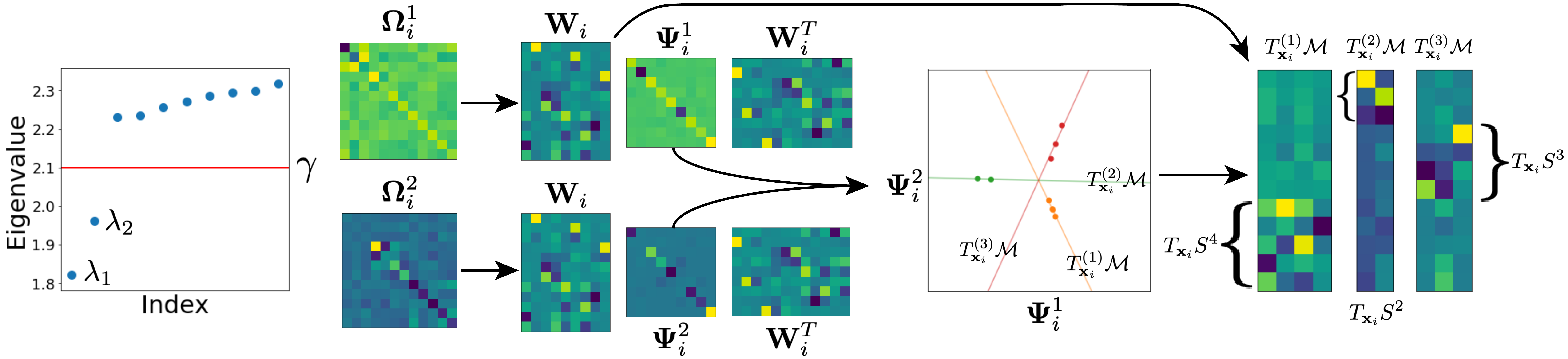}
    \caption{Illustration of the pipeline to go from eigenfunctions of $\laplsym$ to a set of bases for disentangled subspaces, shown on real data from the manifold $S^2 \times S^3 \times S^4$. Matrices $\mathbf{\Omega}^r_i$ up to the threshold in the spectrum $\gamma$ are simultaneously diagonalized as $\mathbf{W}_i \mathbf{\Psi}^r_i \mathbf{W}_i^T$. Columns of $\mathbf{W}_i$ are clustered based on the cosine similarity of the diagonals of $\mathbf{\Psi}^r_i$ to form bases for the subspaces $T^{(1)}_{\mathbf{x}_i}\mathcal{M},\ldots,T^{(m)}_{\mathbf{x}_i}\mathcal{M}$. For clarity, the data is visualized in the embedding space rather than coordinates of the tangent space.}
    \label{fig:pipeline}
    \vspace{-10pt}
\end{figure}

Suppose we have a set of points $\mathbf{x}_1,\ldots,\mathbf{x}_t \in \mathbb{R}^n$ sampled from some manifold embedded in $\mathbb{R}^n$. To construct $\Delta^2$, we first build up approximations to many properties of the manifold. In our discussion here, we will assume the data is embedded in a space where the $\ell_2$ metric in $\mathbb{R}^n$ matches the metric on the manifold. Such an embedding must exist for all manifolds \cite{nash1956imbedding}, but how to learn it is an open question. %When true distances are known but an appropriate embedding is not, \geomancer\, can be extended to use the true distances between points as side information, at the cost of making the algorithm more complex. We will save the discussion of that case for later. \pfau{Maybe........}

To start, we construct a symmetric nearest neighbors graph with edges $\mathcal{E} = \{e_{ij}\}$ between $\mathbf{x}_i$ and $\mathbf{x}_j$. This defines the set of possible steps that can be taken under a random walk. Next, we construct a set of tangent spaces, one per data point, by applying PCA to the difference between $\mathbf{x}_i$ and its neighbors $\mathbf{x}_j$ s.t. $e_{ij}\in \mathcal{E}$. The number of principal components $k$, equivalent to the dimensionality of the manifold, is a hyperparameter we are free to choose. This defines a set of local orthonormal coordinate systems $\mathbf{U}_i$ and local tangent vectors $\mathbf{v}_j$ s.t. $\mathbf{x}_j - \mathbf{x}_i \approx \mathbf{U}_i \mathbf{v}_j$ for neighboring points $\mathbf{x}_j$. This approach to constructing local tangent spaces is also used by many other manifold learning methods \cite{zhang2004principal, singer2011orientability, singer2012vector, salhov2012patch, wolf2013linear}. We will use these coordinates to construct the parallel transport from the point $\mathbf{x}_i$ to $\mathbf{x}_j$.

\paragraph{Graph Connection Laplacians}
To construct $\Delta^2$, we need a generalization of graph Laplacians to higher order tensors. The graph Laplacian is a linear operator on scalar functions on a graph, defined as:

\begin{equation}
    \Delta^0[f]_i = \sum_{j\,\mathrm{s.t.}\,e_{ij}\in\mathcal{E}} f_i - f_j
    \label{eqn:graph_laplacian}
\end{equation}

Equivalently, if we represent functions over data points as a vector in $\mathbb{R}^t$ then the Laplacian can be given as a matrix $\Delta^0$ in $\mathbb{R}^{t \times t}$ with $\Delta^0_{ij} = -1$ if $e_{ij}\in\mathcal{E}$ and $\Delta^0_{ii} = n_i$ where $n_i$ is the number of neighbors of $\mathbf{x}_i$. If the graph is approximating a Riemannian manifold, then in the limit of dense sampling the graph Laplacian becomes equivalent to the Laplace-Beltrami operator \cite{belkin2008towards}.

To generalize the graph Laplacian from scalars to vectors and tensors, we can replace the difference between neighboring scalars in Eqn.~\ref{eqn:graph_laplacian} with a difference between tensors. These must be tensors {\em in the same tangent space}, so any neighboring vectors must be parallel transported from $T_{\mathbf{x}_i}\mathcal{M}$ to $T_{\mathbf{x}_j}\mathcal{M}$, and similarly for higher-order tensors. On a graph, the parallel transport from $\mathbf{x}_i$ to $\mathbf{x}_j$ can be approximated by an orthonormal matrix $\mathbf{Q}_{ij}$ associated with $e_{ij}$, while the transport in the reverse direction is given by $\mathbf{Q}_{ji} = \mathbf{Q}_{ij}^T$. This leads to a natural definition for the first-order {\em graph connection Laplacian} \cite{singer2012vector}:

\begin{equation}
    \Delta^1[\mathbf{v}]_i = \sum_{j\,\mathrm{s.t.}\,e_{ij}\in\mathcal{E}} \mathbf{v}_i - \mathbf{Q}_{ij}^T \mathbf{v}_j
    \label{eqn:graph_connection_laplacian}
\end{equation}
This is a linear operator on vector-valued functions. We can represent vector-valued functions as a single flattened vector in $\mathbb{R}^{tk}$, in which case the graph connection Laplacian is a block-sparse matrix in $\mathbb{R}^{tk \times tk}$. Generalizing to matrices yields the second-order graph connection Laplacian:

\begin{equation}
    \Delta^2[\mathbf{\Sigma}]_i = \sum_{j\,\mathrm{s.t.}\,e_{ij}\in\mathcal{E}} \mathbf{\Sigma}_i - \mathbf{Q}_{ij}^T \mathbf{\Sigma}_j \mathbf{Q}_{ij}
    \label{eqn:gcl-2}
\end{equation}
which, again flattening matrix-valued functions to vectors in $\mathbb{R}^{tk^2}$, gives a block-sparse matrix in $\mathbb{R}^{tk^2 \times tk^2}$. The diagonal blocks $\Delta^2_{(ii)} = n_i \mathbf{I}$ while the block for edge $e_{ij}$ is given by $\Delta^2_{(ij)} = -\mathbf{Q}_{ij} \otimes \mathbf{Q}_{ij}$.

While this gives a general definition of $\Delta^2$ for graphs, we still need to define the connection matrices $\mathbf{Q}_{ij}$. When a manifold inherits its metric from the embedding space, the connection is given by the projection of the connection in the embedding space. As the connection in Euclidean space is trivial, the connection on the manifold is given by the orthonormal matrix that most closely approximates the projection of $T_{\mathbf{x}_j}\mathcal{M}$ onto $T_{\mathbf{x}_i}\mathcal{M}$. In the local coordinates defined by $\mathbf{U}_i$ and  $\mathbf{U}_j$, the projection is given by $\mathbf{U}_j^T\mathbf{U}_i$. If $\mathbf{U}_{ij}\mathbf{\Sigma}_{ij}\mathbf{V}_{ij}^T$ is the SVD of $\mathbf{U}_j^T\mathbf{U}_i$, then $\mathbf{Q}_{ij} = \mathbf{U}_{ij} \mathbf{V}_{ij}^T$ gives the orthonormal matrix nearest to $\mathbf{U}_j^T\mathbf{U}_i$. This is closely related to the canonical (or principal) angles between subspaces \cite{zhu2013angles}, and this connection was also used by Singer and Wu \cite{singer2012vector} for the original graph connection Laplacian.

\begin{algorithm}[t!]
\SetAlgoLined
\KwData{$\mathbf{x}_1, \ldots, \mathbf{x}_t \in \mathbb{R}^n$ sampled from $\mathcal{M} = \mathcal{M}_1 \times \ldots \times \mathcal{M}_m$ with dimension $k$}
\textbf{1. Build the manifold:} \\
 \hspace{.3cm} $e_{ij} \in \mathcal{E}$ if $\mathbf{x}_{j}  \in \mathrm{knn}(\mathbf{x}_i)$ or $\mathbf{x}_{i} \in \mathrm{knn}(\mathbf{x}_j)$ \Comment*[f]{Construct nearest neighbors graph}\\
 \hspace{.3cm} $d\mathbf{X}_i = (\mathbf{x}_{j_1}-\mathbf{x}_i, \ldots, \mathbf{x}_{j_{n_i}}-\mathbf{x}_i)$ for $j_1, \ldots, j_{n_i}$ s.t. $e_{ij}\in\mathcal{E}$\\
\hspace{.3cm} $\mathbf{U}_i \mathbf{\Sigma}_i \mathbf{V}_i^T = \mathrm{SVD}(d\mathbf{X}_i)$, $T_{\mathbf{x}_i}\mathcal{M} \approx \mathrm{span}(\mathbf{U}_i)$ \Comment*[f]{Estimate tangent spaces by local PCA}\\
\textbf{2. Build and diagonalize the connection Laplacian:} \\
\hspace{.3cm} $\mathbf{U}_{ij}\mathbf{\Sigma}_{ij}\mathbf{V}_{ij}^T = \mathrm{SVD}(\mathbf{U}^T_j \mathbf{U}_i)$ \\
\hspace{.3cm} $\mathbf{Q}_{ij} = \mathbf{U}_{ij}\mathbf{V}_{ij}^T$ for all $i, j$ s.t. $e_{ij}\in\mathcal{E}$ \Comment*[f]{Construct connection} \\
\hspace{.3cm} $\Delta^2_{(ij)} = -\mathbf{Q}_{ij}\otimes\mathbf{Q}_{ij}$,\, $\Delta^2_{(ii)} = n_i \mathbf{I}$ \Comment*[f]{Build blocks of 2nd-order graph connection Laplacian} \\
\hspace{.3cm} $\overline{\Delta^2_{(ij)}} = \mathbf{\Pi}_{\mathrm{tr}}^T\mathbf{\Pi}_{\mathrm{sym}}^T\Delta^2_{(ij)}\mathbf{\Pi}_{\mathrm{sym}}\mathbf{\Pi}_{\mathrm{tr}}$ \Comment*[f]{Project blocks onto space of symmetric zero-trace matrices} \\
\hspace{.3cm} $\laplsym${\boldmath$\phi$}$^r = \lambda_r${\boldmath$\phi$}$^r$, $r=1, \ldots, R$ \Comment*[f]{Compute bottom $R$ eigenfunctions/values of $\laplsym$} \\
\hspace{.3cm} $\mathrm{vec}(\mathbf{\Omega}^r_i) = \mathbf{\Pi}_{\mathrm{sym}}\mathbf{\Pi}_{\mathrm{tr}}${\boldmath$\phi$}$^r_i$ \Comment*[f]{Project eigenfunctions back to matrices}\\
\textbf{3. Align the results from different eigenvectors of the Laplacian:} \\
\hspace{.3cm} $\mathbf{W}_i \mathbf{\Psi}^r_i \mathbf{W}_i^T = \mathbf{\Omega}^r_i$ for all $r$ s.t. $\lambda_r < \gamma$ \Comment*[f]{Simultaneously diagonalize matrices by \textsc{FFDiag} {\normalfont \cite{ziehe2004fast}}}\\
%$T^{(\cdot)}_{\mathbf{x}_i}\mathcal{M} = \mathrm{span}(\mathbf{U}_i)$ \Comment*[f]{Initialize disentangled subspaces at $\mathbf{x}_i$}\\
%\While{$\lambda_r < \gamma$}{
%$\mathbf{W}^r_i\mathbf{\Psi}^r_i\mathbf{W}^{rT}_i=\mathbf{\Omega}^r_i$ \Comment*[f]{Take eigendecomposition of $\mathbf{\Omega}^r_i$} \\
%$\mathcal{S}^{r\pm}_i =\{\mathbf{w}^{1}_{ij} | \pm\psi^{1}_{ij} > 0,\, j \in 1, \ldots, k\}$\Comment*[f]{Split tangent space into two orthogonal spaces} \\
%$T^{(j\pm)}_{\mathbf{x}_i}\mathcal{M} = T^{(j)}_{\mathbf{x}_i}\mathcal{M} \cap \mathrm{span}(\mathcal{S}^{r\pm}_i)$ \Comment*[f]{Fold subspaces into existing decomposition}
%}
\hspace{.3cm} {\boldmath$\psi$}$_{ik}=(\Psi^1_{i,kk}\ldots,\Psi^r_{i,kk},\ldots,\Psi^{m-1}_{i,kk})$ \\
\hspace{.3cm} $\mathcal{C}^j = \{${\boldmath$\psi$}$_{ik}| ${\boldmath$\psi$}$_{ik}^T${\boldmath$\psi$} $_{ik'}/||${\boldmath$\psi$}$_{ik}||\,||${\boldmath$\psi$}$_{ik'}||>0.5\}$ \Comment*[f]{Cluster diagonals of $\mathbf{\Psi}_i$ by cosine similarity} \\
\hspace{.3cm} $T^{(j)}_{\mathbf{x}_i}\mathcal{M} = \mathrm{span}(\{\mathbf{w}_{ik}|${\boldmath$\psi$}$_{ik}\in\mathcal{C}^j\})$ \Comment*[f]{Columns of $\mathbf{W}_i$ in each cluster span the subspaces} \\
\KwResult{Orthogonal subspaces $T^{(1)}_{\mathbf{x}_i}\mathcal{M}, \ldots, T^{(m)}_{\mathbf{x}_i}\mathcal{M}$ at every point $\mathbf{x}_i$ tangent to $\mathcal{M}_1, \ldots, \mathcal{M}_m$}
\caption{Geometric Manifold Component Estimator (\geomancer)}
\label{alg:geomancer}
\end{algorithm}

\paragraph{Eliminating Spurious Eigenfunctions}

We now have all the ingredients needed to construct $\Delta^2$. However, a few modifications are necessary to separate out the eigenfunctions that are projections onto submanifolds from those that are due to specific properties of the particular manifold. First, many manifolds have eigenfunctions of $\Delta^2$ with zero eigenvalue that are skew-symmetric (Supp. Mat., Sec.~\ref{sec:antisymmetric_eigenfunctions}). Moreover, the action of $\Delta^2$ on any function of the form $f_j \mathbf{I}$ will be the same as the action of $\Delta^0$ on $f_j$, meaning eigenvalues of $\Delta^0$ are present in the spectrum of $\Delta^2$ as well. While these will not typically have eigenvalue zero, they may still be small enough to get mixed in with more meaningful results. To avoid both of these spurious eigenfunctions we project each block of $\Delta^2$  onto the space of operators on symmetric zero-trace matrices, to yield a projected block $\overline{\Delta^2_{(ij)}}$ of the projected second-order graph connection Laplacian $\laplsym$. The eigenfunctions of interest can still be expressed as $\sum_{j} c^r_j \mathbf{\Pi}^{(j)}$ where $\mathbf{\Pi}^{(j)}_i$ is the orthogonal projection onto $T^{(j)}_{\mathbf{x}_i}\mathcal{M}$ and $\sum_j c^r_j \mathrm{dim}(T^{(j)}_{\mathbf{x}_i}\mathcal{M})=0$. We can then use standard sparse eigensolvers to find the lowest eigenvalues $\lambda_1, \ldots, \lambda_R$ and eigenvectors {\boldmath$\phi$}$^1, \ldots, ${\boldmath$\phi$}$^R$, which we split into individual vectors {\boldmath$\phi$}$^r_i$ for each point $\mathbf{x}_i$ and project back to full $k \times k$ matrices $\mathbf{\Omega}^r_i$. 
For details please refer to Sec.~\ref{sec:eliminating_eigenfunctions_appendix}.

\begin{figure}
    \centering
    \subfloat[][]{
    \includegraphics[width=0.33\textwidth]{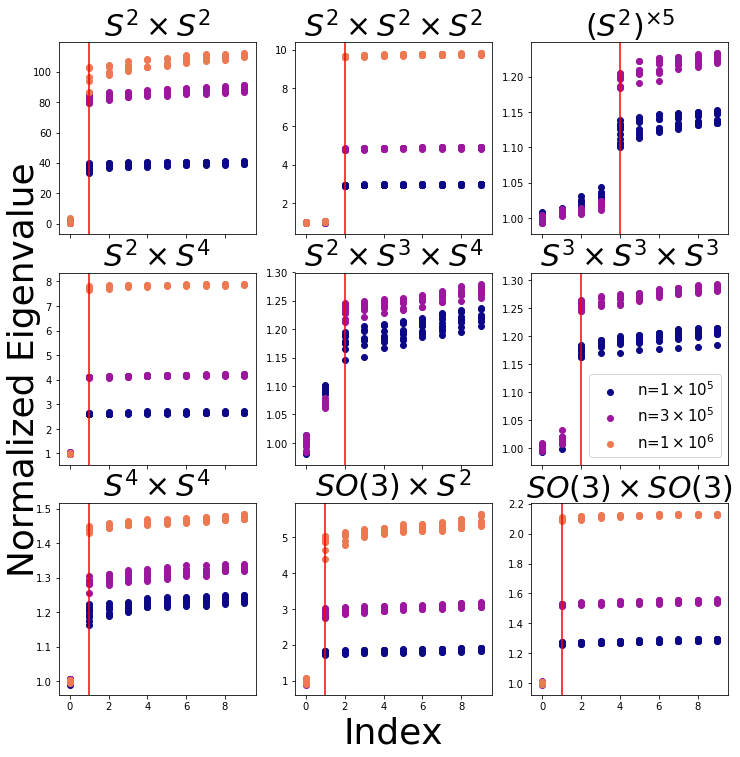}
    }
    \subfloat[][]{
    \includegraphics[width=0.66\textwidth]{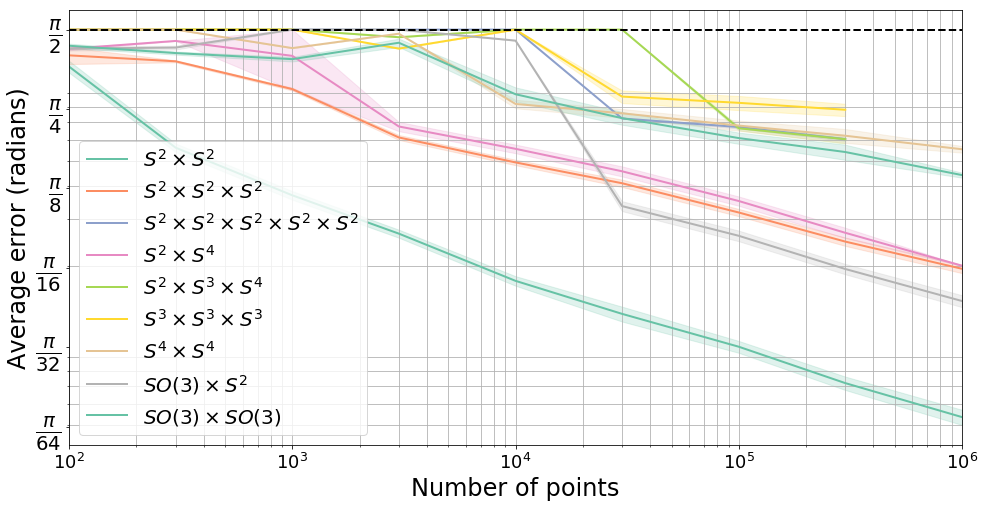}
    }
    \caption{Results on synthetic manifolds. (a) The spectrum of $\laplsym$ for products of spheres and special orthogonal groups with different amounts of data. The spectrum is rescaled so the first eigenvalue equals 1. A clear gap emerges in the spectrum at the eigenvalue equal to the number of submanifolds (red line). (b) The average angle between the subspaces recovered by \geomancer\, and the true tangent spaces of the submanifolds. Past a critical threshold, the error declines with more training data.}
    \label{fig:synthetic_results}
    \vspace{-10pt}
\end{figure}

\paragraph{Clustering Subspace Dimensions}
Once we have computed the smallest eigenvalues $\lambda_1, \ldots, \lambda_R$ and matrices $\mathbf{\Omega}^1_1, \ldots, \mathbf{\Omega}^R_t$ from $\laplsym$, we need to merge the results together into a set of orthogonal subspaces $T^{(1)}_{\mathbf{x}_i}\mathcal{M}, \ldots, T^{(m)}_{\mathbf{x}_i}\mathcal{M}$ at every point. The appropriate number of submanifolds $m$ can be inferred by looking for a gap in the spectrum and stopping before $\lambda_m > \gamma$, similar to how PCA can identify the dimensionality of the best linear projection of a dataset. Due to the degeneracy of the eigenfunctions and the constraint that $\mathrm{tr}(\mathbf{\Omega}^r_i)=0$, the results will be linear combinations of $\mathbf{\Pi}^{(j)}_i$ that we have to demix. As the projection matrices are orthogonal to one another, they can be expressed in the same orthonormal basis $\mathbf{W}_i$ as $\mathbf{\Pi}^{(j)}_i=\mathbf{W}_i \mathbf{D}^{(j)}_i \mathbf{W}^T_i$, where $\mathbf{D}^{(j)}_i$ are $0/1$-valued diagonal matrices such that $\sum_j \mathbf{D}^{(j)}_i = \mathbf{I}$. This is a simultaneous diagonalization problem, which generalizes eigendecomposition to multiple matrices that share the same eigenvectors. We solve this with the orthogonal \textsc{FFDiag} algorithm \cite{ziehe2004fast}, yielding a decomposition $\mathbf{\Omega}^r_i = \mathbf{W}_i \mathbf{\Psi}^r_i \mathbf{W}^T_i$ where $\mathbf{\Psi}^r_i \approx \sum_j c^r_j \mathbf{D}^{(j)}_i$. 

The columns of $\mathbf{W}_i$ then need to be clustered, one cluster per disentangled subspace. Let {\boldmath$\psi$}$_{ik}$ be the vector made up of the $k$-th diagonal of $\mathbf{\Psi}^r_i$ for all $r= 1,\ldots,m-1$. The simultaneous constraints on $c^r_j$ and $\mathbf{D}^{(j)}_i$ push $m-1$-dimensional vectors {\boldmath$\psi$}$_{ik}$ to cluster together in the $m$ corners of a simplex. Thus the vectors can simply be clustered by checking if the cosine similarity between two {\boldmath$\psi$}$_{ik}$ is greater than some threshold. Finally, a basis for every disentangled subspace can be constructed by taking all columns $\mathbf{w}_{ik}$ of $\mathbf{W}_i$ such that {\boldmath$\psi$}$_{ik}$ cluster together. An example is given in Fig.~\ref{fig:pipeline}. %By restricting ourselves to zero-trace eigenfunctions, the matrices $\mathbf{\Omega}^r_i$ will be proportional to linear combinations of projection matrices of the form $\frac{1}{k^r_i}\mathbf{\Pi}^{r+}_i-\frac{1}{k-k^r_i}\mathbf{\Pi}^{r-}_i$, where $k^r_i$ is the dimensionality of the subspace that $\mathbf{\Pi}^{r+}_i$ projects into. A visual interpretation of the meaning of these eigenfunctions is given in Fig.~\ref{fig:diffusion}. We can separate these subspaces by taking the eigendecomposition of $\mathbf{\Omega}^{r}_i = \mathbf{W}^r_i \mathbf{\Psi}^r_i \mathbf{W}^{rT}_i$ at every point $\mathbf{x}_i$ and splitting the eigenvectors $\mathbf{W}^r_i = (\mathbf{w}^r_{i1},\ldots,\mathbf{w}^r_{ik})$ by whether the eigenvalues $\psi^r_{i1}, \ldots, \psi^r_{ik}$ are less than or greater than 0. Denote these sets by $\mathcal{S}_i^{r+}=\{\mathbf{w}^{1}_{ij} | \psi^{1}_{ij} > 0,\, j \in 1, \ldots, k\}$ and $\mathcal{S}_i^{r-}=\{\mathbf{w}^{1}_{ij} | \psi^{1}_{ij} < 0,\, j \in 1, \ldots, k\}$, this gives bases for two disentangled tangent subspaces.

%While this decomposes the tangent space around a data point into two orthogonal subspaces, we may need to look at more than one eigenfunction if there are more than two submanifolds to disentangle. To build up a set of orthogonal subspaces recursively, start with $T^{(1)}_{\mathbf{x}_i}\mathcal{M} = \mathrm{span}(\mathcal{S}^{1+}_i)$ and $T^{(2)}_{\mathbf{x}_i}\mathcal{M} = \mathrm{span}(\mathcal{S}^{1-}_i)$. Then for a given $r$, we can compute the angles between $\mathrm{span}(\mathcal{S}^{r\pm}_i)$ and all $T^{(j)}_{\mathbf{x}_i}\mathcal{M}$ for $j\in 1,\ldots,r$. For most $T^{(j)}_{\mathbf{x}_i}\mathcal{M}$, these angles will either be all 0 or all $\pi/2$, but for the one subspace for which this is not true, we can split the basis into $T^{(j+)}_{\mathbf{x}_i}\mathcal{M}$ and $T^{(j-)}_{\mathbf{x}_i}\mathcal{M}$ depending on which dimensions are in the span of $\mathcal{S}^{r+}_i$. We would expect a gap in the spectrum $\lambda_1, \ldots, \lambda_R$ at the number of true submanifolds, which gives us a convenient signal for when to stop splitting. An example of this pipeline for the product of three spheres $S^2 \times S^3 \times S^4$ is given in Fig.~\ref{fig:pipeline}.

The complete algorithm for \geomancer\, is summarized in Alg.~\ref{alg:geomancer}. The basic building blocks are just nearest neighbors, SVD and eigendecomposition. \geomancer\, requires very few hyperparameters -- just the dimension $k$, the number of nearest neighbors, and the gap $\gamma$ in the spectrum of $\laplsym$ at which to stop splitting tangent spaces, which can be chosen by simple heuristics. We demonstrate \geomancer's performance in the next section.

%When dealing with noisy or incomplete data, exact parallel transport is impossible, and approximations must be used. Instead of invariant subspaces, we look for {\em slowly varying} subspaces. Finally, the metric must be known if we are to construct an affine connection. We can either use the $\ell_2$ metric from the ambient space the data is embedded in, or use a metric provided by some other means. We leave the question of how to discover the metric itself to future work.

\section{Experiments}

% \begin{figure}
%     \centering
%     \includegraphics[width=\textwidth]{bunny_and_dragon_long.png}
%     \caption{The Stanford Bunny and Stanford Dragon under different pose and illumination conditions.}
%     \label{fig:stanford_3d_data}
% \end{figure}

To demonstrate the power of \geomancer, we investigate its performance on both synthetic manifolds and a dataset of rendered 3D objects.\footnote{The code can be found at \texttt{http://tinyurl.com/dm-geomancer}} We avoid using existing performance metrics for disentangling, as most are based on the causal or probabilistic interpretation of disentangling \cite{suter2019robustly, locatello2019challenging, higgins2018towards, kim2018disentangling, chen2018isolating, ridgeway2018learning, eastwood2018framework, duan2020unsupervised} rather than the symmetry-based one. Furthermore, many disentangling metrics assume that each disentangled factor is one-dimensional \cite{kim2018disentangling, chen2018isolating, suter2019robustly, ridgeway2018learning, eastwood2018framework}, which is not the case in our investigation here. Details of dataset generation, training, evaluation and additional results are given in Supp. Mat., Sec.~\ref{sec:experimental_details}.

\paragraph{Synthetic Data}
First, we generated data from a variety of product manifolds by uniformly sampling from either the $n$-dimensional sphere $S^n \subset \mathbb{R}^{n+1}$, represented as vectors with unit length, or the special orthogonal group $SO(n) \subset \mathbb{R}^{n \times n}$, represented as orthonormal matrices with positive determinant. We then concatenated the vectorized data to form a product manifold. As these manifolds are geodesically complete and have a metric structure inherited from their embedding in Euclidean space, we would expect \geomancer\, to discover their global product structure easily.

Due to sampling noise, no eigenvalues were exactly zero, but it can be seen in Fig.~\ref{fig:synthetic_results}(a) that the spectrum has a large gap at the number corresponding to the number of submanifolds, which grows with increased data. In Fig.~\ref{fig:synthetic_results}(b) we show the average angle between the learned subspaces and the ground truth tangent spaces of each submanifold. For all manifolds, a threshold is crossed beyond which the error begins to decline exponentially with more data. Unsurprisingly, the amount of data required grows for more complex high-dimensional manifolds, but in all cases \geomancer\, is able to discover the true product structure. Even for manifolds like $SO(n)$ that are not simply connected, \geomancer\, still learns the correct local coordinate system. Note that the most complex manifolds we are able to disentangle consist of up to 5 submanifolds, significantly outperforming other recent approaches to symmetry-based disentanglement which have not been applied to anything more complex than a product of two submanifolds \cite{casellesdupre2019disentangled,quessard2020learning}.  %toroidal 2D worlds \cite{casellesdupre2019disentangled} or 3D rotation combined with color transformation \cite{quessard2020learning}. 
Moreover, our approach is fully unsupervised, while theirs requires conditioning on actions.

%\begin{wraptable}{r}{10cm}
\begin{table}[t]
    \centering
    \begin{small}
    %\begin{sc}
    \begin{tabular}{c|cccccccc}
          Object & \multicolumn{4}{c}{Latents} & LEM & Pixels & $\beta$-VAE & Chance \\ \cline{2-5}
         &  True & Rotated & Scaled & Linear & d=15 & & d=8 & \\\hline\hline
         Bunny & {\bf 0.024} & {\bf 0.024} & 0.72$\pm$0.68 & 1.42$\pm$0.09 & 0.37 & 1.25 & 1.30$\pm$0.02 & 1.26$\pm$0.23 \\
         Dragon & {\bf 0.023} & {\bf 0.023} & 0.96$\pm$0.70 & 1.26$\pm$0.37 & 0.32 & 1.26 & 1.15$\pm$0.09 & 1.26$\pm$0.23 \\
    \end{tabular}
    %\end{sc}
    \end{small}
    \caption{The average angle between the true disentangled subspaces and the subspaces recovered by \geomancer\, from different embeddings of Stanford 3D objects. The first 5 columns all use information from the true latents, including Laplacian Eigenmaps (LEM) with 15 embedding dimensions, while the next 2 columns only use pixels.}
    \label{tab:stanford_3d_results}
    \vspace{-20pt}
\end{table}

%\end{wraptable}

\paragraph{Stanford 3D Objects}

\begin{wrapfigure}{r}{0.3\textwidth}
    \vspace{-10pt}
    \centering
    \includegraphics[width=0.3\textwidth]{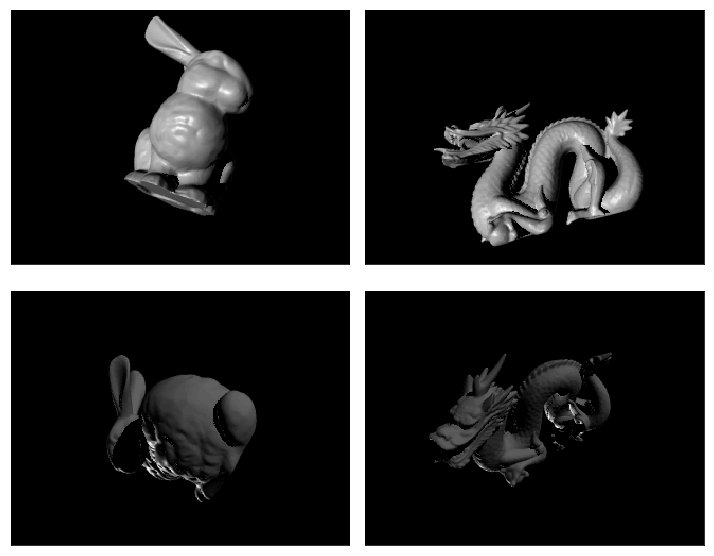}
    \caption{The Stanford Bunny and Dragon with different poses and illumination.}
    \label{fig:stanford_3d_data}
    \vspace{-10pt}
\end{wrapfigure}

To investigate more realistic data, we applied \geomancer\, to renderings of the Bunny and Dragon object from the Stanford 3D Scanning Repository \cite{levoy2005stanford} under different pose and lighting conditions (Fig.~\ref{fig:stanford_3d_data}).\footnote{The data can be found at \texttt{http://tinyurl.com/dm-s3o4d}} We chose to render our own data, as existing datasets for 3D objects are limited to changes in azimuth and a limited range of elevations \cite{lecun2004learning, liu2015faceattributes, fidler20123d, kim2018disentangling, aubry2014seeing, paysan20093d}. Instead, we sampled rotations of object pose uniformly from $SO(3)$, while the light location was sampled uniformly from the sphere $S^2$. In Table~\ref{tab:stanford_3d_results}, we show the accuracy of \geomancer\, applied to several different embeddings using the same performance metric as in Fig.~\ref{fig:synthetic_results}(b). When applied directly to the true latent state vectors, \geomancer\, performs exceptionally well, even if the state is rotated randomly. When individual dimensions are multiplied by a random scale factor, the performance degrades, and if a random linear transformation is applied to the state, performance is no better than chance. This shows that accurate metric information is necessary for \geomancer\, to work. We also applied our method to embeddings learned by Laplacian Eigenmaps using no information other than knowledge of the nearest neighbors in the true latent space. While not as accurate as working from the true latent state, it still performs far better than chance, showing that metric information alone is {\em sufficient} for \geomancer\, to work.

Trying to disentangle directly from pixels is more complicated. As the mapping from state to observation is highly nonlinear, \geomancer\, performs no better than chance directly from pixels.\todopfau{If we do not get results in time, replace "\geomancer\, performs" with "we expect \geomancer\, would perform"} However, existing algorithms to disentangle directly from pixels fail as well \cite{higgins2017beta} (See Supp. Mat., Sec.~\ref{sec:beta_vae_training}). Even when applying \geomancer\, to the latent vectors learned by the $\beta$-VAE, the results are no better than chance. The poor performance of both \geomancer\, and $\beta$-VAE on the Stanford 3D Objects shows that disentangling full 3D rotations from pixels without side information remains an open problem.

\section{Discussion}
We have shown that \geomancer\, is capable of factorizing manifolds directly from unstructured samples. On images of 3D objects under changes in pose and illumination, we show that correct metric information is critical for \geomancer\, to work. There are several directions for improvement. As \geomancer\, is a nonparametric spectral method, it does not automatically generalize to held-out data, though there are extensions that enable this \cite{bengio2004out, pfau2019spectral}. While \geomancer\, scales well with the amount of data, the number of nonzero elements in $\laplsym$ grows as $\mathcal{O}(k^4)$ in the dimensionality of the manifold, meaning new approximations are needed to scale to more complex manifolds. The motivating mathematical intuition could also be implemented in a parametric model, enabling end-to-end training. 

The missing ingredient for fully unsupervised disentangling is a source of correct metric information. We have shown that existing generative models for disentangling are insufficient for learning an embedding with the right metric structure. We hope that this will be a challenge taken up by the metric learning community. We are hopeful that this charts a new path forward for disentangling research.

\section*{Broader Impact}

The present work is primarily theoretical, making its broader impact difficult to ascertain. We consider the algorithm presented here to be a potential core machine learning method, and as such it could have an impact in any area that machine learning can be applied to, but particularly in unsupervised learning, computer vision and robotic manipulation.

\begin{ack}
We would like to thank  M{\'e}lanie Rey, Drew Jaegle, Pedro Ortega, Peter Toth, Olivier H{\'e}naff, Malcolm Reynolds, and Kevin Musgrave for helpful discussions and Shakir Mohamed for support and encouragement.
\end{ack}

\bibliography{geomancer_arxiv_final}

%\newpage
\appendix

%\input{christoffel_notes.tex}

%\suptitle

\section{Review of Differential Geometry}
\label{sec:diff_geom_review}

\subsection{Riemannian manifolds}

Consider a $k$-dimensional manifold $\mathcal{M}$. At every point $x\in\mathcal{M}$, the tangent space $T_{x}\mathcal{M}$ is a $k$-dimensional vector space made up of all velocity vectors $\dot{\gamma}(t)$ where $\gamma:\mathbb{R}\rightarrow\mathcal{M}$ is a path such that $\gamma(t)=x$. There are many different ways that a manifold can be embedded in a vector space (for instance, the manifold of natural images can be embedded in the vector space of pixel representations of an image), and quantities on the manifold must be defined in a way that they transform consistently between different embeddings. Let $\mathbf{x}\in \mathbb{R}^n$ be an embedding of the point $x$. Under a differentiable change in embedding $\overline{\mathbf{x}} = f(\mathbf{x})$, tangent vector components $\mathbf{v}$ transform as $\overline{\mathbf{v}} = \mathbf{J}_f \mathbf{v}$, where $\mathbf{J}_f$ is the Jacobian of $f$ at $\mathbf{x}$. The cotangent space $T_x^*\mathcal{M}$ is also a $k$-dimensional vector space, but it consists of all gradients of differentiable functions at $x$ and for finite dimensional manifolds is the dual space to the tangent space. A cotangent vector $\mathbf{w}$ transforms under a change of coordinates as $\mathbf{w} = \mathbf{J}_f^{-1} \mathbf{w}$. When not otherwise specified, we will use ``vector" and ``tangent vector" interchangeably. Spaces of higher-order tensors can be defined based on how they transform under changes of coordinates. For instance, a linear transform of vectors in $T_x\mathcal{M}$ represented by the matrix $\mathbf{A}$ transforms as $\overline{\mathbf{A}} = \mathbf{J}_f \mathbf{A} \mathbf{J}_f^{-1}$, so linear transforms are rank-(1,1) tensors in $T_x\mathcal{M} \otimes T_x^*\mathcal{M}$.

In a Riemannian manifold, every point $x$ is equipped with a metric $\langle \cdot, \cdot \rangle_x: T_x\mathcal{M}\times T_x\mathcal{M}\rightarrow \mathbb{R}$ that defines distances locally. If we choose a basis for the tangent space $T_x\mathcal{M}$, then in that basis the metric can be represented as a positive definite matrix $\mathbf{G}_x\in \mathbb{S}^n_+$ and $\langle \mathbf{v}, \mathbf{w}\rangle_x = \mathbf{v}^T\mathbf{G}_x\mathbf{w}$. This includes the $\ell_2$ metric as a special case when $\mathbf{G}_x = \mathbf{I}$, and has the same form as the Mahalanobis distance from statistics \citep{mahalanobis1936generalized}, but for tangent vectors instead of distributions. Critically, the metric can change when moving across the manifold. The metric transforms as $\mathbf{G}_{\overline{x}} = \mathbf{J}_f^{-T}\mathbf{G}_x\mathbf{J}_f^{-1}$, so the metric is a rank-(0,2) tensor in $T^*_x\mathcal{M} \otimes T^*_x\mathcal{M} = T^{* \otimes 2}_x\mathcal{M}$. For cotangent vectors, the metric is $\langle \mathbf{v}, \mathbf{w} \rangle^*_x = \mathbf{v}^T \mathbf{G}^{-1}_x \mathbf{w}$, which is a rank-(2,0) tensor.

Once the metric is known in a given coordinate system, the Laplace-Beltrami operator can also be constructed in terms of coordinates:

\begin{equation}
    \Delta[f](x) = \frac{1}{\sqrt{\mathrm{det}(\mathbf{G}_x)}}\sum_j \frac{\partial}{\partial x_j}\left(\sqrt{\mathrm{det}(\mathbf{G}_x)}\sum_i g^{-1}_{ij} \frac{\partial f}{\partial x_i}\right)
    \label{eqn:laplace-beltrami}
\end{equation}

In flat Euclidean space, $\mathbf{G}_{x} = \mathbf{I}$ and this reduces to the more familiar Laplacian $\Delta[f] = \sum_i \frac{\partial^2 f}{\partial x_i^2}$.

For any two points $x, y\in\mathcal{M}$, the geodesic distance between them is defined as the minimum length of any path between them
\begin{equation}
    \mathcal{D}(x, y) = \min_{\substack{\gamma \\ \gamma(0)=x \\ \gamma(1)=y}} \int_0^1 dt \sqrt{\langle \dot{\gamma}(t), \dot{\gamma}(t) \rangle_{\gamma(t)}}
\end{equation}
A {\em geodesic} between $x$ and $y$ is a locally shortest path that is parameterised by arc length. In other words it is a path such that there exists a constant $c$ with: $\forall t\in[0,1),\exists\epsilon>0|~\forall t'\in[0,\epsilon] \mathcal{D}(\gamma(t), \gamma(t')) = c(t' - t)$. Note that a geodesic is not necessarily a minimum path from start to end. For example, a great circle from the south pole to itself on a sphere is a geodesic even though the distance from the south pole to itself is of course $0$.

It's worth noting that the metric is a purely {\em local} notion of distance, defined only in the tangent space, while the geodestic distance is a {\em global} distance between two points anywhere on the manifold. Despite the name, the term ``metric learning" in machine learning typically refers to learning a single, global notion of distance, or to learning a mapping that {\em preserves} distances, under the assumption that the correct local distance is already known \cite{kulis2013metric}.

\subsection{Parallel transport and affine connections}

So far we described how to construct a vector space equipped with a metric at every point on the manifold, but have not given any way to relate vectors in one tangent space to those in another. In general there is not a unique mapping from vectors in one tangent space to another, which is precisely why the usual parallelogram model of analogy breaks down when dealing with curved manifolds. Instead, a vector in $T_x\mathcal{M}$ can be identified with a vector in $T_y\mathcal{M}$ in a path-dependent manner through a process called {\em parallel transport}, where the vector is moved infinitesimally along a path such that it is always {\em locally} parallel with itself as it moves. To do this, we have to define what it means to be ``locally parallel", which requires additional machinery: the {\em affine connection}.

The affine connection at $x$ is a map $\Gamma_x: T_x\mathcal{M}\times T_x\mathcal{M}\rightarrow T_x\mathcal{M}$. For two vectors $\mathbf{v}$ and $\mathbf{w}\in T_x\mathcal{M}$, $\Gamma_x(\mathbf{v}, \mathbf{w})$ can be intuitively thought of as the amount the vector $\mathbf{v}$ changes when moving to an infinitesimally nearby tangent space in the direction $\mathbf{w}$. For a Riemannian manifold, there are two natural properties that an affine connection should obey: it should preserve the metric, which means that the inner product between vectors does not change when they are parallel transported, and it should be torsion-free, which intuitively means the vector should not ``twist" as it is parallel-transported. Given the appropriate formal definition of these requirements, there is a unique connection that satisfies these properties: the {\em Levi-Civita} connection. For a given choice of coordinates such that the metric can be represented by $\mathbf{G}_x$ at $x$, and letting the $ij$th element of $\mathbf{G}_x$ be denoted $g_{ij}$ and the $i$th element of $\Gamma_{x}(\mathbf{v}, \mathbf{w})$ be denoted $\Gamma_x (\mathbf{v}, \mathbf{w})_i$, the Levi-Civita connection at $x$ can be written in terms of the Christoffel symbols

\begin{align}
    \Gamma_x (\mathbf{v}, \mathbf{w})_i &= \sum_{jk} \Gamma^i_{jk} v_j w_k \nonumber\\
    \Gamma^i_{jk} &= \frac{1}{2}\sum_\ell g^{-1}_{i\ell}\left(\frac{\partial g_{\ell k}}{\partial x_j} + \frac{\partial g_{\ell j}}{\partial x_k} - \frac{\partial g_{jk}}{\partial x_\ell}\right)
\end{align}

The Levi-Civita connection defines a {\em covariant derivative} which takes a vector field $\mathbf{v}:\mathcal{M}\rightarrow T_{x}\mathcal{M}$ and a direction $\mathbf{w}\in T_x\mathcal{M}$ and gives the derivative of the field in that direction $\nabla_{\mathbf{w}}\mathbf{v}(x) = \frac{\partial \mathbf{v}}{\partial \mathbf{w}}\vert_x + \Gamma_x(\mathbf{v}(x), \mathbf{w})$. For a manifold embedded in $\mathbb{R}^n$ that also inherits the metric from this space, the covariant derivative is the ordinary derivative in $\mathbb{R}^n$ plus a correction to keep the vector on the manifold, where the affine connection is precisely that correction. In other words, the covariant derivative is the projection of the ordinary derivative onto the manifold. For other Riemannian manifolds, it is better thought of as a correction to force the covariant derivative to transform correctly as a rank-(0,1) tensor. It's worth noting that, as the Levi-Civita connection is a {\em correction} to make the covariant derivative transform correctly, the connection itself does {\em not} transform as a tensor. For a change of coordinates $x \rightarrow \overline{x}$, the Christoffel symbols transform as:

\begin{equation}
    \overline{\Gamma}^i_{jk} = \sum_{mnp} \frac{\partial \overline{x}_i}{\partial x_m}\left[\Gamma^m_{np}\frac{\partial x_n}{\partial \overline{x}_j}\frac{\partial x_p}{\partial \overline{x}_k} + \frac{\partial^2 x_m}{\partial\overline{x}_j \partial \overline{x}_k}\right]
\end{equation}
where the first term in the sum is the usual change of coordinates for a rank-(1,2) tensor, and the second term is the correction to account for the change in curvature.

Informally, two vectors can be thought of as parallel if the covariant derivative in the direction from one to the other is zero. That is, for some infinitesimal $dt$ and vectors $\mathbf{v}, \mathbf{w}\in T_x\mathcal{M}$, the vector $\mathbf{v}+\Gamma_x(\mathbf{v}, \mathbf{w})dt$ in the tangent space of $x + \mathbf{w} dt$ will be parallel to $\mathbf{v}$. Formally, for a path $\gamma:[0, 1]\rightarrow \mathcal{M}$, the parallel transport of the starting vector $\mathbf{v}(0)\in T_{\gamma(0)}\mathcal{M}$ is a function $\mathbf{v}(t)\in T_{\gamma(t)}\mathcal{M}$ such that $\nabla_{\dot{\gamma}(t)}\mathbf{v}(t)=\dot{\mathbf{v}}(t)+\Gamma_{\gamma(t)}\left(\mathbf{v}(t), \dot{\gamma}(t)\right)=0$. Parallel transport  makes it possible to define a differential equation to solve for the geodesic: if the velocity vector of a path is a parallel transport, that is, if $\nabla_{\dot{\gamma}(t)}\dot{\gamma}(t) =\ddot{\gamma}(t)+\Gamma_{\gamma(t)}\left(\dot{\gamma}(t), \dot{\gamma}(t)\right)=0$, then $\gamma$ is a geodesic. An intuitive way to think of this is that a geodesic is a path that always goes ``straight forward" locally -- its acceleration is always parallel to the path.

Parallel transport can also be defined for higher-order tensors. For a rank-$(p,q)$ tensor $a_{i_1,\ldots,i_p}^{i^*_1,\ldots,i^*_q}\in T_x^{\otimes p} \otimes T_x^{*\otimes q}$, the differential equation that defines the parallel transport is given by contracting the Christoffel symbols over all $r$ indices of the tensor:

\begin{equation}
    \frac{\partial a_{i_1,\ldots,i_p}^{i^*_1,\ldots,i^*_q}}{\partial t} =  \sum_{j_1 \ldots j_p, j^*_1 \ldots j^*_q, k} \Gamma^{i_1}_{j_1 k} \ldots \Gamma^{i_p}_{j_p k} \Gamma^{j^*_1}_{i^*_1 k} \ldots \Gamma^{j^*_q}_{i^*_q k} a_{j_1,\ldots,j_p}^{j^*_1,\ldots,j^*_q}(t) \dot{\gamma}_k(t) 
    \label{eqn:tensor_parallel_transport}
\end{equation}

\section{Definitions and Proofs}
\label{sec:proofs}

%\begin{theorem*}{The Levi-Civita Connection of an Embedded Submanifold.\footnote{This proof is taken from a discussion on Math StackExchange: \texttt{https://tinyurl.com/yaj3bbkm}.}}
%Let $\mathcal{M}$ be a Riemannian manifold with metric $\langle \cdot, \cdot \rangle_{\mathcal{M}}$ and $\mathcal{N} \subset \mathcal{M}$ be a submanifold that inherits the metric from $\mathcal{M}$, so that for $\mathbf{v}$, $\mathbf{w}\in T_x \mathcal{N}\subset T_x \mathcal{M}$, $\langle \mathbf{v}, \mathbf{w} \rangle_{\mathcal{N}} = \langle \mathbf{v}, \mathbf{w} \rangle_{\mathcal{M}}$. Then the Levi-Civita connection on $\mathcal{N}$ is given by the projection of the Levi-Civita connection on $\mathcal{M}$ into the tangent space on $\mathcal{N}$. Let $\nabla_\cdot \cdot$ denote the connection on $\mathcal{M}$ and $\Tilde{\nabla}_\cdot \cdot$ denote the connection on $\mathcal{N}$. That means, for two vector fields $X$ and $Y$ whose derivatives at any point $x$ on $\mathcal{N}$ are restricted to be in $T_x\mathcal{N}$:
%\[
%\Tilde{\nabla}_X Y = \mathbf{\Pi}_{\mathcal{N}} \nabla_X Y
%\]
%where $\mathbf{\Pi}_{\mathcal{N}}$ is the orthogonal projection of a vector in $T_x\mathcal{M}$ onto $T_x\mathcal{N}$.
%\end{theorem*}

%\begin{proof}
%The Levi-Civita connection is the unique connection that is torsion-free and compatible under the metric. Compatibility under the metric is obvious from the fact that the projection onto $T_x\mathcal{N}$ leaves vectors already in $T_x\mathcal{N}$ invariant, and the fact that the metric on $\mathcal{N}$ is inherited from the metric on $\mathcal{M}$.
%\end{proof}

\theoremstyle{definition}
\begin{defn}{Groups, actions and orbits:}
A {\em group} is a set $G = \{g, h, \ldots\}$ equipped with a composition operator $\cdot: G\rightarrow G$ such that:
\begin{enumerate}
    \item $G$ is closed under composition: $g\cdot h \in G$ $\forall g,\, h \in G$
    \item There exists an identity element $e\in G$ such that $g \cdot e = e \cdot g = g$ $\forall g \in G$
    \item The composition operator is associative: $f \cdot (g \cdot h) = (f \cdot g) \cdot h \,\forall f,\,g,\,h\in G$
    \item For all $g\in G$, there exists an inverse element $g^{-1} \in G$ such that $g\cdot g^{-1} = g^{-1} \cdot g = e$
\end{enumerate}
For some other object $z \in Z$, a group {\em action} is a function $\cdot:G\times Z\rightarrow Z$ s.t. $e \cdot z = z$ and $(gh) \cdot z = g \cdot (h \cdot z)\, \forall z\in Z$ and $g, h \in G$. The set $Z$ of all objects under the action of all group elements is referred to as the {\em orbit} of $z$ under the action of $G$. For instance, the unit sphere is the orbit of a unit vector under all rotations.
\end{defn}
\begin{defn}{Symmetry-Based Disentangling (Higgins et al., 2018):}
Let $W$ be the set of world states, $G$ be a group that acts on those world states which factorizes as $G=G_1\times G_2\times \ldots \times G_m$, and $f:W\rightarrow Z$ be a mapping to a latent representation space $Z$. The representation $Z$ is said to be {\em disentangled} with respect to the group factorization $G=G_1\times G_2\times \ldots \times G_m$ if:
\begin{enumerate}
    \item There exists an action of $G$ on $Z$.
    \item The map $f:W\rightarrow Z$ is equivariant between the actions of $G$ on $W$ and $Z$, i.e. $g \cdot f(w) = f(g \cdot w)~\forall g \in G, w \in W$, and
    \item There is a fixed decomposition $Z = Z_1 \times Z_2 \times \ldots \times Z_m$ such that each $Z_i$ is invariant to the action of $G_j$ for all $j$ except $j=i$.
\end{enumerate}
\end{defn}
\vspace{0.5cm}

\begin{theorem*}{Main text, Theorem 2:}
Let $\mathcal{M} = \mathcal{M}_1 \times \ldots \times \mathcal{M}_m$ be a Riemannian product manifold, and let $T^{(1)}_x\mathcal{M},\ldots,T^{(m)}_x\mathcal{M}$ denote orthogonal subspaces of $T_x\mathcal{M}$ that are tangent to each submanifold. Then the tensor fields $\mathbf{\Pi}^{(i)}:\mathcal{M}\rightarrow T_x\mathcal{M}\otimes T^*_x\mathcal{M}$ for $i \in 1, \ldots, m$, where $\mathbf{\Pi}^{(i)}_x$ is the linear projection operator from $T_x\mathcal{M}\rightarrow T^{(i)}_x\mathcal{M}$, are in the kernel of the connection Laplacian $\Delta^2$.
\end{theorem*}
\begin{argument}
Given a basis $\mathbf{U}^{(i)}_x$ of the subspace $T^{(i)}_x$, the projection matrix is given by $\mathbf{\Pi}^{(i)}_x = \mathbf{U}^{(i)}_x\mathbf{U}^{(i)T}_x$. As $T^{(i)}_x$ is an invariant subspace under parallel transport, the holonomy of $\mathbf{U}^{(i)}_x$ around any loop has the form $\mathbf{U}^{(i)}_x\mathbf{Q}$ for some orthonormal matrix $\mathbf{Q}$. Therefore the holonomy of $\mathbf{\Pi}^{(i)}_x$ is given by $\mathbf{U}^{(i)}_x\mathbf{Q}\mathbf{Q}^T\mathbf{U}^{(i)T} = \mathbf{U}^{(i)}_x\mathbf{U}^{(i)T} = \mathbf{\Pi}^{(i)}_x$, and $\mathbf{\Pi}^{(i)}_x$ is invariant to parallel transport. As the rate of change for the tensor field $\mathbf{\Pi}^{(i)}$ under diffusion will be 0, the entire tensor field goes to 0 under the action of $\Delta^2$.
\end{argument}

\begin{proof}
Each $\mathbf{\Pi}^{(i)}$ is an endomorphism of the tangent bundle. For a general endomorphism $u$ of the tangent bundle, and a general vector field $X$, the covariant derivative satisfies $\nabla (u(X)) = (\nabla u)(X) + u(\nabla X)$. Let's replace $u$ with $\mathbf{\Pi}^{(i)}$ in this formula. Since $\mathcal{M}$ is a product of Riemannian manifold, we have that $\nabla\left(\mathbf{\Pi}^{(i)}(X)\right)$ and $\mathbf{\Pi}^{(i)}(\nabla X)$ are equal. It follows that $\nabla\mathbf{\Pi}^{(i)}$ is always $0$, and the Laplacian $\Delta^2 \mathbf{\Pi}^{(i)} = \mathrm{Tr}\nabla^2 \mathbf{\Pi}^{(i)}$ also has to be $0$.
\end{proof}

\section{Spurious Eigenfunctions of $\Delta^2$}
\label{sec:antisymmetric_eigenfunctions}

\subsection{Analysis}

\begin{figure}[h!]
    \centering
    \includegraphics[width=\textwidth]{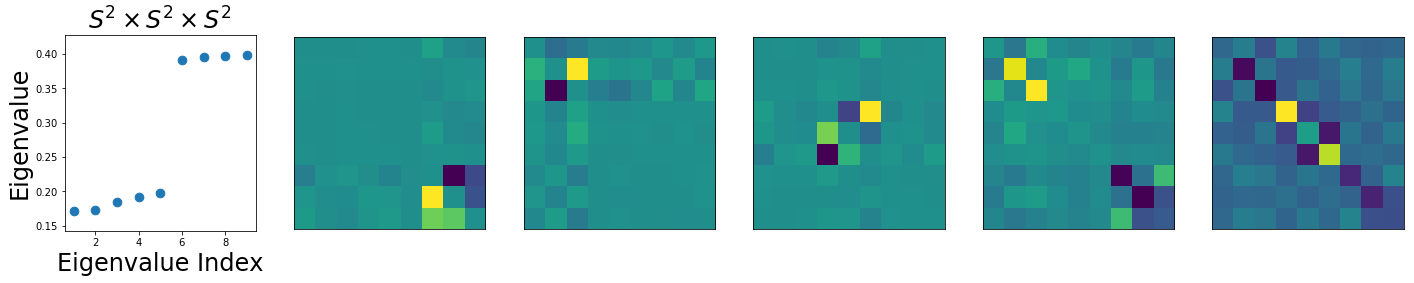}
    \caption{Eigenvalues and eigenfunctions of $\Delta^2$ for the product manifold $S^2 \times S^2 \times S^2$, without restriction to symmetric matrices. The spectrum (left) clearly has 5 nontrivial but small values before the gap. The value of the first 5 nontrivial eigenfunctions at a single point are shown in the remaining figures. The first three are clearly the skew-symmetric volume form, while the remaining two are the expected projection matrices.}
    \label{fig:antisymmetric_eigenfunctions_s2}
\end{figure}

\begin{figure}[h!]
    \centering
    \includegraphics[width=\textwidth]{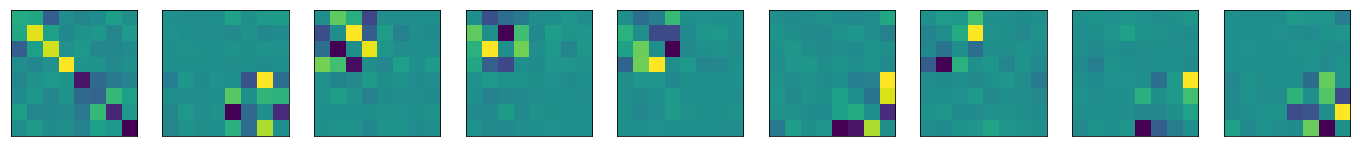}
    \caption{Eigenfunctions of $\Delta^2$ for the product manifold $S^3 \times S^3$, without restriction to symmetric matrices. The first nontrivial eigenfunction is the expected projection matrix, while the next eight eigenfunctions are all skew-symmetric -- four per manifold.}
    \label{fig:antisymmetric_eigenfunctions_s3}
\end{figure}

A complete characterization of the zero eigenfunctions of the second-order connection Laplacian is beyond the scope of this paper. However, we have both empirically and theoretically found several zero eigenfunctions not of the form of projection matrices onto factor manifolds. The spheres $S^2$ and $S^3$ in particular seem to have a zero eigenfunction which maps points on the manifold to a {\em skew-symmetric} matrix.

In Figs.~\ref{fig:antisymmetric_eigenfunctions_s2}~and~\ref{fig:antisymmetric_eigenfunctions_s3}, we give examples of these eigenfunctions at a random point on $(S^2)^{\times 3}$ and $(S^3)^{\times 2}$. Each submanifold of $S^2$ has a single skew-symmetric eigenfunction, while each submanifold of $S^3$ has four such skew-symmetric eigenfunctions. Looking at the spectrum, the eigenvalues are of similar magnitude to those eigenvalues used by \geomancer. Indeed, the individual eigenfunctions separate the submanifolds of interest so cleanly that it is unfortunate that these eigenfunctions do not seem to exist for all manifolds.

For $S^2$ we can construct the skew-symmetric eigenfunction as follows. Let $(\mathbf{v}_1, \mathbf{v}_2)$ be an orthonormal basis for a point on $S^2$, then $\mathbf{v}_1\mathbf{v}_2^T - \mathbf{v}_2\mathbf{v}_1^T$ is a skew-symmetric tensor. This tensor does not depend on the choice of basis, so it is a uniquely defined tensor field on the whole of $S^2$ (This is in fact one way to construct the volume form for $S^2$). As parallel transport preserves orthonormality, this field is left invariant by parallel transport. Any field which is invariant under parallel transport is a zero eigenfunction of the connection Laplacian. For general spheres $S^n$, the volume form will be a rank-$n$ skew-symmetric tensor, and therefore will not in general be an eigenfunction of $\Delta^2$. The interpretation of the 4 skew-symmetric zero eigenfunctions that exist for $S^3$ is still an open question, and we also do not know whether these skew-symmetric eigenfunctions exist for other manifolds. \todoseb{Discuss $S^3$ and the trivial decomposition of the tangent bundle?}

\subsection{Eliminating Spurious Eigenfunctions}
\label{sec:eliminating_eigenfunctions_appendix}

To remove skew-symmetric eigenfunctions, let $\mathbf{\Pi}_{\mathrm{sym}}$ be the linear projection operator from $R^{k \times k}$ to the space of symmetric matrices, which can be represented by a matrix in $\mathbb{R}^{k^2 \times k(k+1)/2}$. Then we can project the blocks of $\Delta^2$ into this smaller space to remove eigenfunctions which are skew-symmetric. Note that this is a projection of full matrices into a lower dimensional space where the matrix is only represented by its upper (or lower) triangular, rather than a projection into the space of full matrices. This has the added benefit of reducing the computational overhead in both space and time by about a factor of 4.

To avoid eigenfunctions derived from $\Delta^0$, which will always have the form of some scalar function times the identity, we further multiply the blocks by $\mathbf{\Pi}_{\mathrm{tr}}\in\mathbb{R}^{k(k+1)/2 \times k(k+1)/2 - 1}$, which projects symmetric matrices onto symmetric matrices with zero trace. Putting this all together, we project each block $\Delta^2_{(ij)}$ onto the space of operators on symmetric zero-trace matrices, to yield a projected block $\overline{\Delta^2_{(ij)}} = \mathbf{\Pi}_{\mathrm{tr}}^T\mathbf{\Pi}_{\mathrm{sym}}^T\Delta^2_{(ij)}\mathbf{\Pi}_{\mathrm{sym}}\mathbf{\Pi}_{\mathrm{tr}}$ and projected second-order graph connection Laplacian $\laplsym$.

\section{Experimental Details}
\label{sec:experimental_details}

For all experiments, we used twice the dimensionality of the manifold for the number of nearest neighbors, and computed the bottom 10 eigenfunctions of $\laplsym$. We chose the threshold $\gamma$ such that the algorithm would terminate at the largest gap in the spectrum. We ran 10 copies of all experiments to validate the robustness of our results. All experiments were run on CPU. The simplest experiments finished within minutes (for instance, $S^2 \times S^2$ with 10,000 data points) while the most complex manifolds required days. The largest experiments, such as $(S^2)^{\times 5}$ with 1,000,000 data points, were terminated after 5 days. On the Stanford 3D Objects data, we implemented some steps in parallel across 100-1000 CPUs, such as computing tangent spaces or connection matrices. This allowed us to complete most steps in \geomancer\, in just a few minutes.

\subsection{Synthetic Manifolds}

\label{sec:additional_synthetic_analysis}
\begin{figure}
    \centering
    \includegraphics[width=\textwidth]{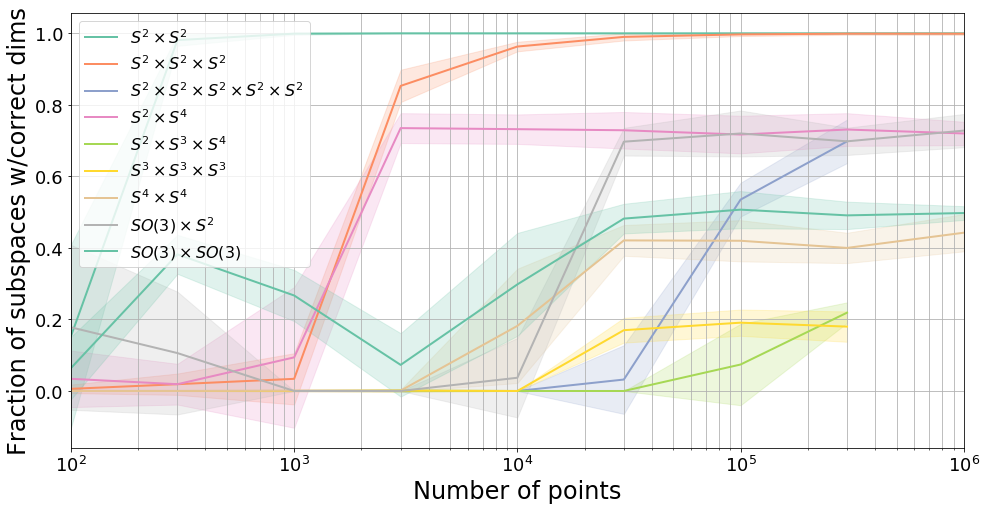}
    \caption{Fraction of points in the training set for which the number and dimensionality of the disentangled subspaces is correctly recovered for synthetic products of spheres and special orthogonal groups. Beyond a critical threshold, the fraction quickly jumps up and plateaus, and on some small manifolds it reaches nearly perfect accuracy.}
    \label{fig:dimensionality_accuracy}
\end{figure}

For the results in Fig.~\ref{fig:synthetic_results}(b), we excluded points where the shape of the subspaces was not estimated correctly. In Fig.~\ref{fig:dimensionality_accuracy}, we count the proportion of points in the training set for which we recovered the correct number and dimensionality of subspaces  and find that, past a threshold in the dataset size, the fraction of correct subspace shapes jumps up, and in some cases becomes essentially exact. The fraction of estimated subspaces with the correct shape and the error between those subspaces and the ground truth seem to rise in tandem, which suggests that there is a hard lower limit on the amount of data required for disentangling.

If $\theta_{i,jk}\in(0, \frac{\pi}{2})$ is the largest angle between the ground truth $T^{(j)}_{\mathbf{x}_i}\mathcal{M}$ and the \geomancer\, estimate of $T^{(k)}_{\mathbf{x}_i}\mathcal{M}$ (or $\frac{\pi}{2}$ if the dimensions do not match), then the error in Fig.~\ref{fig:synthetic_results}(b) is given as $\frac{1}{t}\sum_{i=1}^t \min_{\sigma\in S_m}  \frac{1}{m}\sum_{j=1}^m \theta_{i,j\sigma_j}$ where the minimum is taken over all permutations of $m$ subspaces. 

\subsection{Stanford 3D Objects}

\begin{figure}
    \centering
    \includegraphics[width=\textwidth]{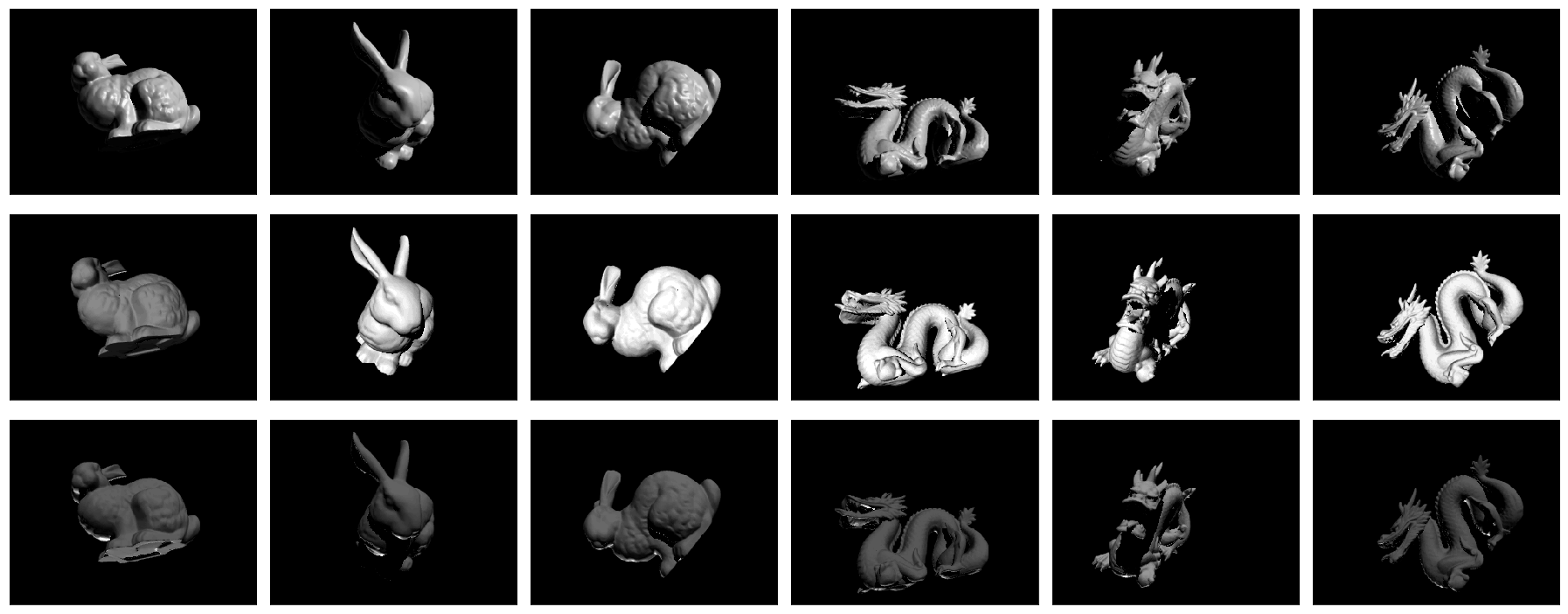}
    \caption{Additional example renderings of the Stanford Bunny and Stanford Dragon under different pose and illumination conditions.}
    \label{fig:bunny_and_dragon}
\end{figure}

A dataset of 100,000 images each of the Stanford Bunny and Stanford Dragon was rendered in MuJoCo \cite{todorov2012mujoco}, originally at 1024x1024 resolution, and downsampled to 64x64 pixels. Images were rendered with a randomly sampled 3D rotation and a randomly sampled illumination source position on a sphere. Latent vectors were represented by a concatenation of unit vectors in $\mathbb{R}^3$ and orthogonal matrices in $\mathbb{R}^{3 \times 3}$ for a 12-dimensional state vector.

When applying \geomancer\, to data other than the true latent state vectors, we can no longer directly compare against ground truth. Instead, we must align the subspaces around the ground truth data with the subspaces around the training data. Let $\mathbf{z}_1, \ldots \mathbf{z}_t$ be the true latent state vectors and $\mathbf{x}_1, \ldots, \mathbf{x}_t$ be the training data. For each point $\mathbf{x}_i$, we use the basis for the tangent space $\mathbf{U}_{\mathbf{x}_i}$ computed in \geomancer, while the tangent space basis $\mathbf{U}_{\mathbf{z}_i}$ for $\mathbf{z}_i$ can be computed in closed form because we know the ground truth is $S^2 \times SO(3)$. Let $i_1, \ldots, i_k$ be the indices of the nearest neighbors of $\mathbf{z}_i$, then we project $\mathbf{z}_{i_1}, \ldots, \mathbf{z}_{i_k}$ into the basis $\mathbf{U}_{\mathbf{z}_i}$ and $\mathbf{x}_{i_1}, \ldots, \mathbf{x}_{i_k}$ into the basis $\mathbf{U}_{\mathbf{x}_i}$ to form a data matrices $V_{\mathbf{z}_i} = \mathbf{U}_{\mathbf{z}_i}^T\left(\mathbf{z}_{i_1}, \ldots, \mathbf{z}_{i_k}\right)$ and $V_{\mathbf{x}_i} = \mathbf{U}_{\mathbf{x}_i}^T\left(\mathbf{x}_{i_1}, \ldots, \mathbf{x}_{i_k}\right)$. We can then align the two subspaces by computing the orthonormal matrix closest to $\left(V_{\mathbf{z}_i}^T V_{\mathbf{z}_i}\right)^{-1/2}V_{\mathbf{z}_i}^T V_{\mathbf{x}_i}\left(V_{\mathbf{x}_i}^T V_{\mathbf{x}_i}\right)^{-1/2}$ using the same SVD technique used to compute the connection matrices in \geomancer. We then compute the angle between ground truth subspaces and subspaces learned by \geomancer\, {\em after} multiplying by the alignment matrix to give the results in Table.~\ref{tab:stanford_3d_results}.

\begin{figure}
    \centering
    \includegraphics[width=\textwidth]{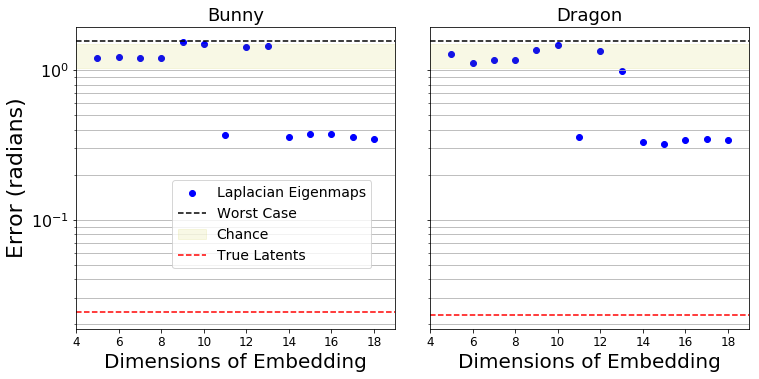}
    \caption{Results of \geomancer\, trained on embedding from Laplacian Eigenmaps (LEM) with different embedding dimensionalities, using the nearest neighbors from the true latents. While not as accurate as working from the true latents directly, \geomancer\, on LEM embeddings performs significantly better than chance above a certain number of dimensions.}
    \label{fig:lem_results}
\end{figure}

The different perturbations applied to the data in Table~\ref{tab:stanford_3d_results} were random orthogonal rotations (Rotated), multiplication by a diagonal matrix with entries sampled from $\mathrm{exp}(\mathcal{N}(0, 0.5))$ (Scaled), and multiplication by a random matrix with entries sampled iid from $\mathcal{N}(0, 1)$ (Linear). For Laplacian Eigenmaps, two points were considered neighbors if the state vector of one was in the 10 nearest neighbors of the other. Varying numbers of embedding dimensions were used, from 5 to 18, and used as input to \geomancer\, (Fig.~\ref{fig:lem_results}). Above 13 dimensions, \geomancer\, consistently performs significantly better than chance.

\subsection{Training $\beta$-VAE on Stanford 3D Objects}
\label{sec:beta_vae_training}
\paragraph{Model architecture}
We used the standard architecture and optimization parameters introduced in\cite{higgins2017beta} for training the $\beta$-VAE model on the Stanford Bunny and Stanford Dragon datasets. The encoder consisted of four convolutional layers (32x4x4 stride 2, 32x4x4 stride 2, 32x4x4 stride 2, and 32x4x4 stride 2), followed by a 128-d fully connected layer and a 32-d latent representation. The decoder architecture was the reverse of the encoder. We used ReLU activations throughout. The decoder parametrized a Bernoulli distribution. We used Adam optimizer with $1e-4$ learning rate and trained the models for 1~mln iterations using batch size of 16, which was enough to achieve convergence. The models were trained to optimize the following disentangling objective:

\begin{equation}
    \mathcal{L}_{\ensuremath{\beta}-VAE} = \mathbb{E}_{p(\mathbf{x})} [\ \mathbb{E}_{q_{\phi}(\mathbf{z}|\mathbf{x})} [\log\ p_{\theta}(\mathbf{x} | \mathbf{z})] - \beta KL(q_{\phi}(\mathbf{z}|\mathbf{x})\ ||\ p(\mathbf{z}))\ ]
\end{equation}

where $p(\mathbf{x})$ is the probability of the image data, $q(\mathbf{z}|\mathbf{x})$ is the learned posterior over the latent units given the data, and $p(\mathbf{z})$ is the unit Gaussian prior with a diagonal covariance matrix. For each dataset we trained 130 instances of the $\beta$-VAE with different $\beta$ hyperparameter sampled uniformly from $\beta \in [1, 30]$ and ten seeds per $\beta$ setting. 

\begin{figure}
    \centering
    \includegraphics[width=\textwidth]{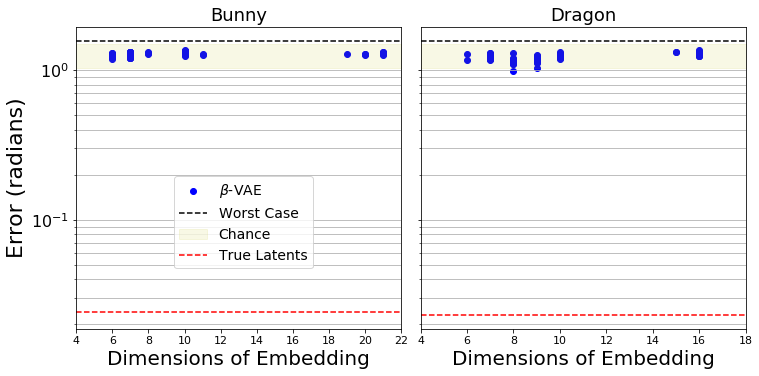}
    \caption{Results of \geomancer\, trained on embedding from $\beta$-VAE with different embedding dimensionalities induced by different values of $\beta$. The $\beta$-VAEs themselves were trained directly from pixels with no knowledge of the true latents. The results are no better than chance.}
    \label{fig:beta_vae_results}
\end{figure}

\paragraph{Model selection}
In order to analyze whether any of the trained $\beta$-VAE instances were able to disentangle the two generative subspaces (changes in 3D rotation and lighting), we applied the recently proposed Unsupervised Disentanglement Ranking (UDR) score \cite{duan2020unsupervised} that measures the quality of disentanglement achieved by trained $\beta$-VAE models by performing pairwise comparisons between the representations learned by models trained using the same hyperparameter setting but with different seeds. This approach requires no access to the ground truth data generative process, and does not make other limiting assumptions that precluded us from applying any other existing disentangling metrics. We used the Spearman version of the UDR score. For each trained $\beta$-VAE model we performed 9 pairwise comparisons with all other models trained with the same $\beta$ value and calculated the corresponding UDR$_{ij}$ score, where $i$ and $j$ index the two $\beta$-VAE models. Each UDR$_{ij}$ score is calculated by computing the similarity matrix $R_{ij}$, where each entry is the Spearman correlation between the responses of individual latent units of the two models. The absolute value of the similarity matrix is then taken $|R_{ij}|$ and the final score for each pair of models is calculated according to:  

\begin{equation}
\label{rel_strength}
\frac{1}{d_a+d_b}\left [  \sum_b  \frac{r_a^2 * I_{KL}(b)}{\sum_a R (a, b)}  +  \sum_a \frac{r_b^2 * I_{KL}(a)}{\sum_b R (a, b)} \right ]  
\end{equation}

where $a$ and $b$ index into the latent units of models $i$ and $j$ respectively, $r_a = \max_a R (a, b)$ and $r_b = \max_b R (a, b)$. $I_{KL}$ indicate the ``informative'' latent units within each model, and $d$ is the number of such latent units. The final score for model $i$ is calculated by taking the median of UDR$_{ij}$ across all $j$.

\begin{figure}[ht]
    \centering
    \includegraphics[width=\textwidth]{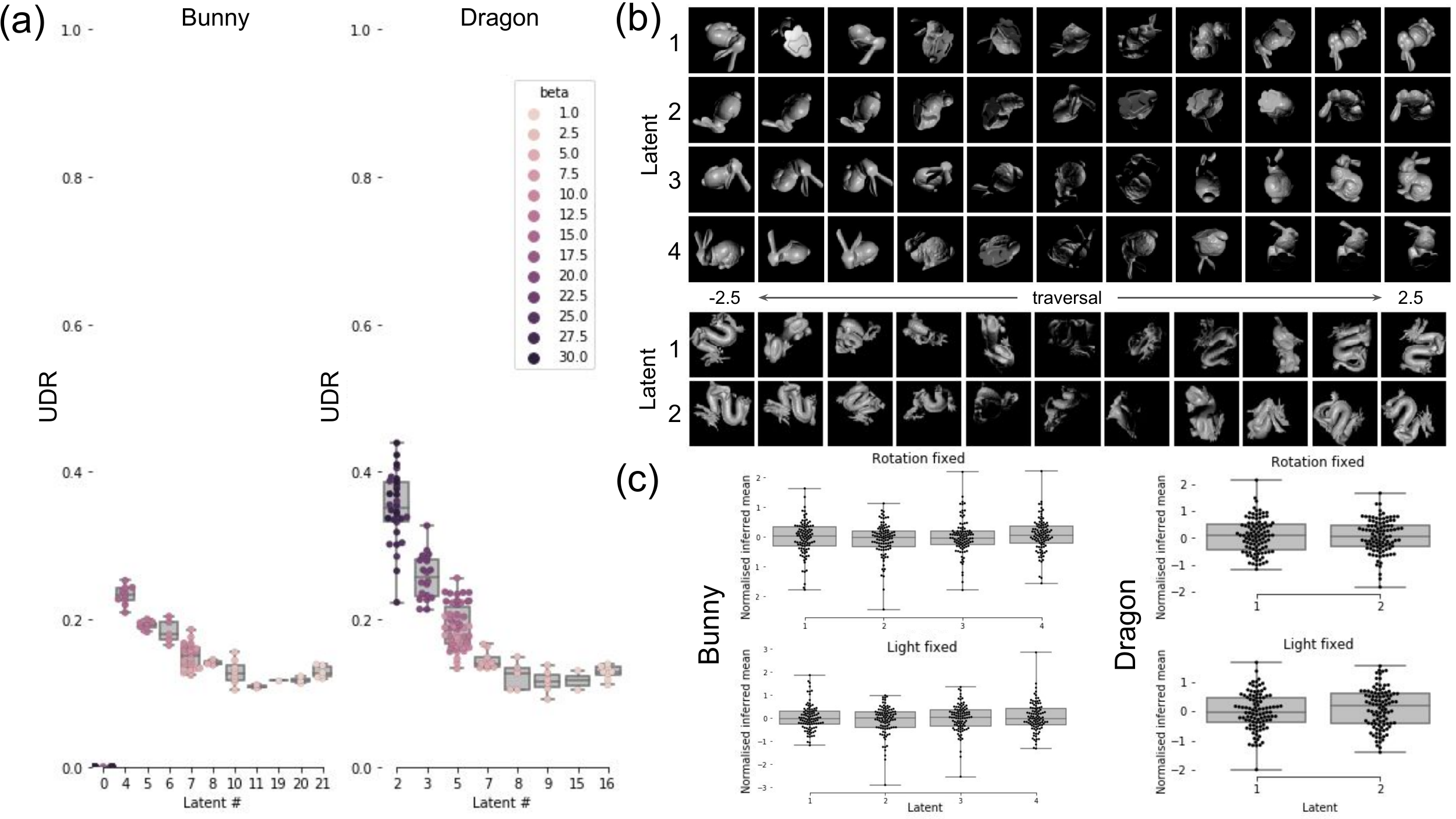}
    \caption{\textbf{(a)} Unsupervised Disentanglement Ranking (UDR) scores \cite{duan2020unsupervised} for 130 $\beta$-VAE models trained with different $\beta$ hyperparameter settings, with ten seeds per setting. UDR scores are plotted against the number of informative latents discovered by the trained model. \textbf{(b)} Latent traversals for the $\beta$-VAE models with the highest UDR scores from (a). An initial set of values for the latents is inferred from a seed image, before changing the value of each latent dimension between -2.5 and 2.5 in equal increments and visualizing the corresponding reconstruction. All latents encode changes in both rotation and lighting. \textbf{(c)} Inferred means for the informative latents from the $\beta$-VAE models with the highest UDR scores from (a). In each subplot 100 images are presented to the model where the value of one subspace (lighting or rotation) is fixed, while the value of the other subspace is randomly sampled. The plotted inferred means are normalized according to ($\mu_i - \overline{\mu}$), where $\overline{\mu}$ is the mean over 100 inferred means $\mu_i$ for the model latent $i$. If a model learns to disentangle lighting from rotation, then latent dimensions corresponding to each disentangled subspace should show significantly smaller dispersion of inferred means in the condition where the corresponding subspace is fixed. It can be seen that no such latents exist in either of the two $\beta$-VAEs.
}
    \label{fig:geomancer_beta_vae_results}
\end{figure}

\paragraph{$\beta$-VAE is unable to disentangle Stanford 3D Objects}
Fig.~\ref{fig:geomancer_beta_vae_results}(a) shows plots UDR scores for the 130 trained $\beta$-VAE models. It is clear that the range of $\beta$ values explored through the hyperparameter search is adequate, since the highest value of $\beta=30$ resulted in the total collapse of the latent space to the prior (resulting in 0 informative latents for the bunny dataset), and the lowest value of $\beta=1$ resulted in too many informative latents to represent SO(3)xS2 in a disentangled manner. None of the trained models were able to achieve high UDR scores close to the maximum of 1. The highest UDR scores were achieved by the models with either two or four informative latents, so we visualized whether they were able to learn a disentangled latent representation that factorizes into independent subspaces representing changes in pose and illumination. Fig.~\ref{fig:geomancer_beta_vae_results}(b) shows that this is not the case, since manipulations of every latent results in changes in both the position and illumination of the Stanford objects. As a final test we presented the same two models with sets of 100 images of the respective Stanford objects that they were trained on. In each set we fixed the value of one of the object's attributes (pose or illumination), while randomly sampling the other one. A model that is able to disentangle these attributes into independent subspaces should have informative latent dimensions with small variance in their inferred means in the condition where their preferred attribute is fixed. It is clear that no such latents exist for the two $beta$-VAE models, with all informative latents encoding both pose and lighting attributes.   

\todopfau{Maybe we can add an example of what happens if we misspecify the dimension?}

\end{document}